\documentclass[]{fairmeta}

\usepackage{tabularx}
\usepackage{array}
\usepackage{multirow}
\usepackage{url}
\usepackage{xcolor}

\usepackage{amsmath,amsfonts,bm}

\def\eqref#1{equation~\ref{#1}}

\def\1{\bm{1}}

\DeclareMathAlphabet{\mathsfit}{\encodingdefault}{\sfdefault}{m}{sl}
\SetMathAlphabet{\mathsfit}{bold}{\encodingdefault}{\sfdefault}{bx}{n}

\usepackage{float}
\usepackage{adjustbox}
\usepackage{eso-pic}
\usepackage{needspace}
\usepackage{enumitem}
\usepackage{flafter}
\setlist[itemize]{topsep=5pt,itemsep=5pt,parsep=0pt}
\titlespacing*{\section}{0pt}{15pt plus 2pt minus 1pt}{7pt}
\titlespacing*{\subsection}{0pt}{11pt plus 2pt minus 1pt}{5pt}
\titlespacing*{\subsubsection}{0pt}{9pt plus 1pt minus 1pt}{4pt}
\pretocmd{\section}{\FloatBarrier\Needspace*{5\baselineskip}}{}{}
\pretocmd{\subsection}{\Needspace*{4\baselineskip}}{}{}
\newenvironment{figuretextblock}{\par\noindent\begin{minipage}{\linewidth}}{\end{minipage}\par}

\newcommand{\LayoutFit}[1]{\adjustbox{max width=\linewidth}{#1}}
\fancypagestyle{firstheader}{\fancyhf{}}
\AtBeginDocument{%
  \AddToShipoutPictureFG*{%
    \AtPageUpperLeft{%
      \put(0,-54){\makebox[\paperwidth][c]{\rule{\textwidth}{0.75pt}}}%
      \put(0,-46){\makebox[\paperwidth][r]{%
        \includegraphics[height=36pt]{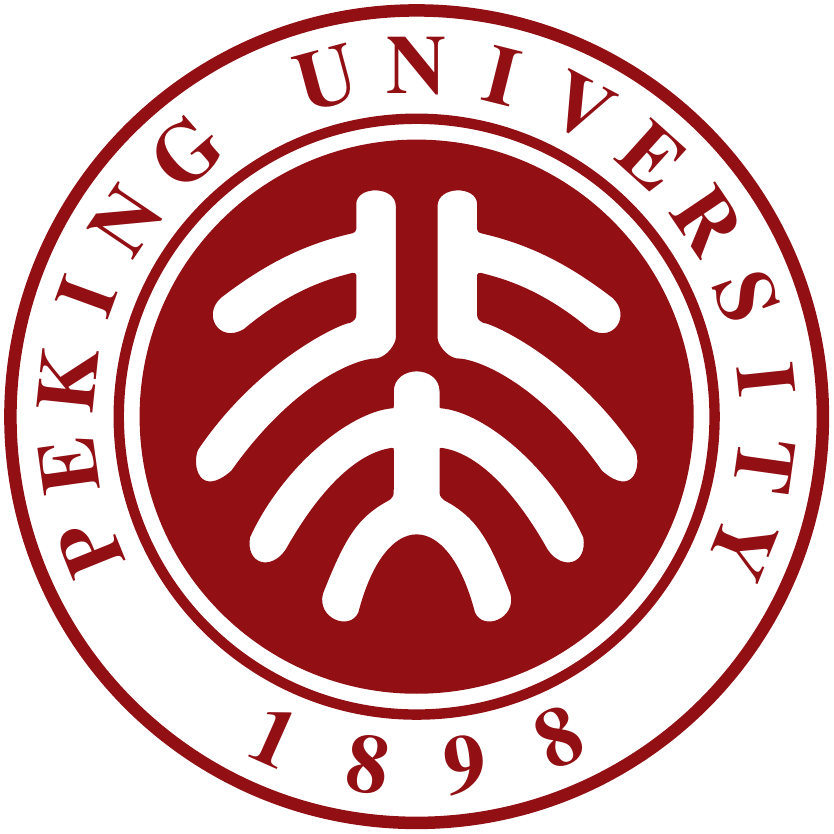}\hspace*{2.5cm}}}%
    }%
  }%
}

\hypersetup{
  colorlinks=true,
  linkcolor=metabg,
  citecolor=metabg,
  urlcolor=metabg,
  pdftitle={Cracking Mobile Manipulation: From Failure Diagnosis to Agentic Robot System},
  pdfauthor={Zixuan Wang, Jia-Ning Li, Yinzhou Tang, Yanzhao Yu, Biao Wu, Meng Fang, \\ Xin Li, Yiwu Zhong}
}

\title{Cracking Mobile Manipulation: From Failure Diagnosis to Agentic Robot System}

\author[1,2,3,4,*]{Zixuan Wang}
\author[5]{Jia-Ning Li}
\author[3]{Yinzhou Tang}
\author[5]{Yanzhao Yu}
\author[6]{Biao Wu}
\author[7]{Meng Fang}
\author[3]{Xin Li}
\author[1,2,\dagger]{Yiwu Zhong}

\affiliation[1]{School of Intelligence Science and Technology, Peking University,}
\affiliation[2]{State Key Laboratory of General Artificial Intelligence, Peking University,}
\affiliation[3]{Alibaba DAMO Academy, Alibaba Group,}
\affiliation[4]{Institute of Automation, Chinese Academy of Sciences,}
\affiliation[5]{Tsinghua University,}
\affiliation[6]{University of Technology Sydney,}
\affiliation[7]{University of Liverpool}

\makeatletter
\renewcommand\contribution[2][]{\addtolist[#1]{#2}{\contributionlist}{\contributionformat}{\quad}}
\makeatother

\contribution[*]{Work done during internship at Alibaba Group.}
\contribution[\dagger]{Corresponding author.}

\let\fairmetaSetAbstract\abstract
\renewenvironment{abstract}{}{}
\fairmetaSetAbstract{\begin{abstract}

Recent vision-language-action (VLA) models have shown strong performance on local robot manipulation, yet their success does not readily translate to long-horizon mobile manipulation. In our experiments, a pretrained subtask VLA achieves \textbf{80\%} success rate on atomic manipulation tasks but only \textbf{19\%} when these skills are composed into mobile manipulation. 
We argue that this gap stems from failures across the execution process rather than from local manipulation capability alone. 
To uncover these bottlenecks, we systematically diagnose failures in long-horizon mobile manipulation and identify three dominant regimes: task planning, navigation and base positioning, and local manipulation. 
Based on this diagnosis, we develop failure-guided agentic intervention that augments a fixed subtask VLA with demonstration-based planning guidance, adaptive visual memory, accessibility-aware base adjustment, and local end-effector adjustment. 
On RoboCasa365, these interventions progressively improve task success from 19\% to \textbf{47\%}, with the finalized system reaching \textbf{49\%} on held-out episode seeds, compared with 18\% for the baseline. 
The same design improves performance from 14\% to \textbf{37\%} on BEHAVIOR-1K and from 5/20 to \textbf{13/20} on real-robot tasks. 
These results suggest that reliable mobile manipulation can be advanced by diagnosing and intervening on the current bottleneck, rather than uniformly scaling the underlying policy. Our code will be released at \url{https://github.com/PKU-VaLuE-Lab/Cracking-Mobile-Manipulation.git}.

\end{abstract}

}

\begin{document}
\newcommand{\proj}{\textsc{MobileAgent}}
\newcommand{\NR}{\textemdash}
\newcommand{\tool}[1]{\texttt{#1}}
\newcommand{\pending}[1]{\textit{[To report: #1]}}
\newcolumntype{Y}{>{\raggedright\arraybackslash}X}

\newcommand{\fixme}[1]{{\textcolor{blue}{\textit{#1}}}}
\newcommand{\tofixme}[1]{{\textcolor{red}{\textit{#1}}}}

\thispagestyle{firstheader}
\maketitle
\pagestyle{empty}

\section{Introduction}

Recent vision-language-action (VLA) models have achieved strong performance on robot manipulation, particularly for the tasks that can be completed within a local tabletop \citep{brohan2023rt2,black2024pi0,black2025pi05,wen2025dexvla}. Agentic systems further extend these capabilities to long-horizon mobile manipulation, through high-level planning, execution feedback, and tool use \citep{ichter2023saycan,zhang2026harnessvla,fu2026capx}. 
Yet mobile manipulation exposes a striking gap between solving individual subtasks and reliably composing them over time. Unlike tabletop manipulation, a mobile robot is required to move across workspaces while maintaining task progress, and every movement changes the viewpoint and base configuration from which the next manipulation is executed \citep{yenamandra2023homerobot,fu2024mobilealoha,liu2024okrobot,wu2025momanipvla}. As a result, strong local manipulation do not necessarily translate into reliable long-horizon mobile manipulation. 

\begin{figure}[!t]
    \centering
    \includegraphics[width=0.95\linewidth]{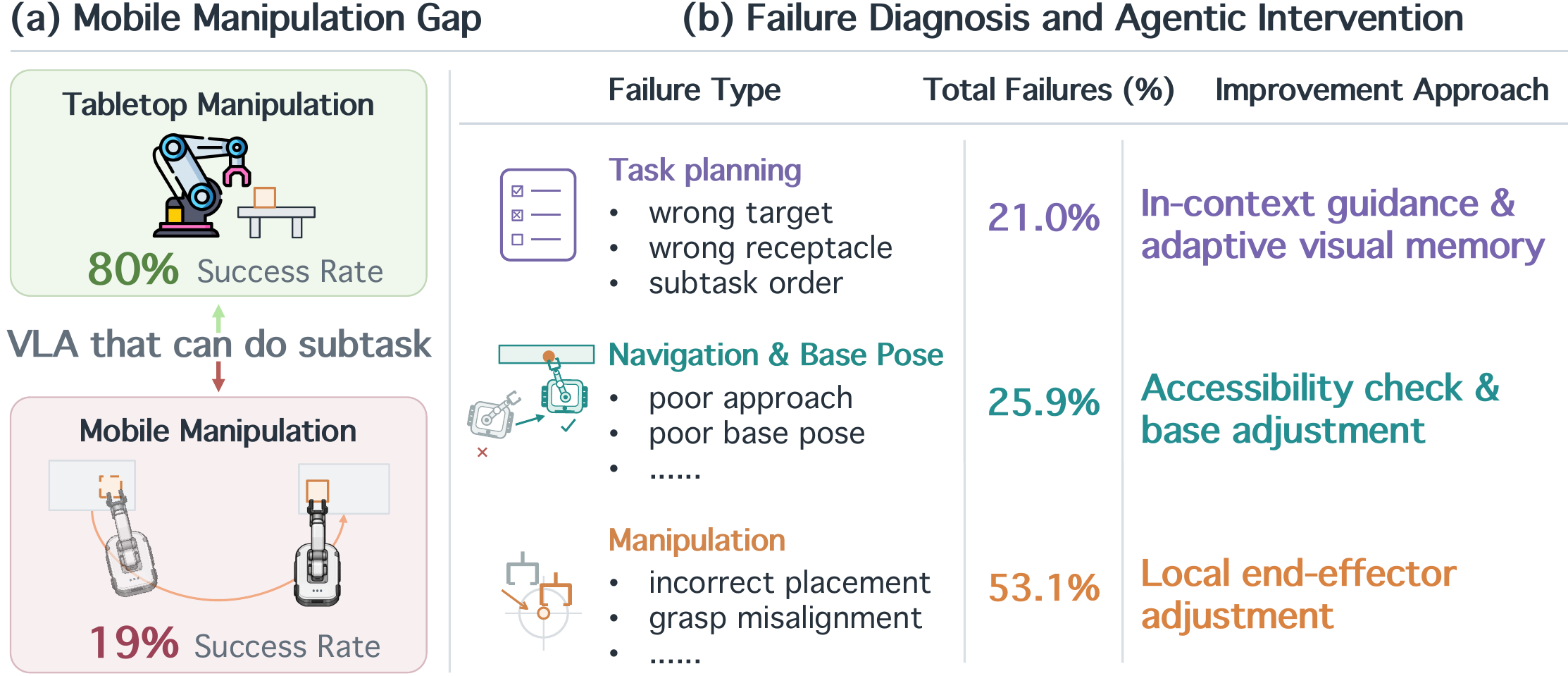}
    
    \caption{ \textbf{From the mobile manipulation gap to failure-guided intervention.} (a) Strong subtask VLA does not translate to reliable long-horizon mobile manipulation. (b) Failure analysis identifies the dominant failures: task planning, navigation and base positioning, and manipulation. These observations motivate our agentic intervention, including: in-context guidance from video demonstrations, adaptive visual memory, accessibility checking and base adjustment, and end-effector adjustment.
}
    \label{fig:figure1}
    
\end{figure}

The gap is surprisingly large. As shown in Figure~\ref{fig:figure1}, a pretrained subtask VLA achieves \textbf{80\%} success rate on atomic manipulation tasks, yet the corresponding agentic baseline achieves only \textbf{19\%} on long-horizon mobile manipulation tasks. 
This discrepancy suggests that the central challenge is not simply whether the robot knows how to manipulate, but whether it can reliably carry a manipulation skill through the sequence of decisions and physical states required by mobile manipulation tasks. 
Specifically, a long-horizon task can fail even before manipulation begins: the planner may select the wrong target or lose track of task progress; navigation may reach the correct workspace but leave the robot in an unusable base pose; and even from a seemingly appropriate configuration, small errors in localization or end-effector alignment can derail the final interaction~\citep{yenamandra2023homerobot,liu2024okrobot,huang2023voxposer}. 
In this sense, what reliable mobile manipulation requires is not simply making a local skill stronger, but identifying where the execution pipeline breaks down.

This leads to a fundamental question: \textbf{\textit{Where does mobile manipulation actually fail over long-horizon tasks, and how can an agent intervene at the bottleneck?}} We answer this question by cracking the failure structure of mobile manipulation. 
Starting from a pretrained VLA that already performs individual tabletop subtasks well, we systematically analyze its failures when these subtasks are composed together. Our analysis reveals three dominant failure regimes that form a natural progression through the execution: task planning, navigation and base positioning, and local manipulation.

Planning errors prevent the robot from pursuing the correct subtask, while navigation errors prevent it from reaching an executable configuration. And manipulation errors remain after the robot has successfully reached the intended workspace. 
This diagnosis motivates a simple but powerful principle: intervene where the current system breaks, rather than uniformly increasing the capability of the underlying policy.

Guided by this principle, we develop a failure-guided agentic framework that progressively adds targeted interventions around a fixed subtask VLA. 
For planning, in-context guidance from video demonstrations provides explicit task procedures~\citep{zhong2023procedure,wang2023mimicplay}, while adaptive visual memory retains what has already been accomplished and what remains to complete~\citep{yoneda2024statler,sridhar2026memer}. 
For navigation, an accessibility check evaluates whether the current robot configuration is suitable for the upcoming manipulation, followed by base adjustment when the approach is geometrically inadequate. 
Finally, for local execution, end-effector adjustment refines the interaction pose to compensate for residual spatial misalignment. 
These components are not independent add-ons: each is directly motivated by a diagnosed failure mode, transforming failure analysis into targeted agentic intervention.

We evaluate the resulting agentic system on all mobile manipulation composite tasks in RoboCasa365 \citep{nasiriany2026robocasa365}. Despite 80\% success rates on atomic tasks, the agentic baseline achieves only 19\% task success in the long-horizon tasks. 
Progressively introducing the proposed interventions raises success rate to \textbf{47\%}. 

This remarkable gain generalizes to held-out episode seeds, improving baseline from 18\% to \textbf{49\%} success rate.
The same design also improves performance beyond RoboCasa365, raising the task success rate from 14\% to \textbf{37\%} on five BEHAVIOR-1K tasks \citep{li2023behavior} and from 5/20 to \textbf{13/20} on two real-robot tasks. 
Taken together, these results suggest a different way to approach long-horizon mobile manipulation. \textbf{Rather than treating the robot as a monolithic policy whose capability must be scaled uniformly, we view it as an agentic system whose reliability can be improved by diagnosing bottleneck and intervening it precisely.} 

\Needspace*{6\baselineskip}
Our contributions are summarized as follows:

\begin{itemize}

\item \textbf{A systematic failure diagnosis of mobile manipulation.} We analyze the gap between strong subtask-level manipulation and long-horizon mobile manipulation, identifying three dominant failure regimes: task planning, navigation and base positioning, and local manipulation. The analysis turns overall success gap into a concrete, stage-wise failure structure.

\item \textbf{A failure-guided agentic intervention framework.} Based on the diagnosed bottlenecks, we introduce targeted interventions around a fixed subtask VLA, including demonstration-based planning guidance, adaptive visual memory, accessibility-aware base adjustment, and local end-effector adjustment.

\item \textbf{Agentic intervention consistently improves the performance of mobile manipulation.} The proposed interventions improve the success rate from 19.0\% to 47.0\% on RoboCasa365. Our experiments further demonstrate consistent gains on held-out RoboCasa365 episodes, BEHAVIOR-1K tasks, and a real dual-arm mobile manipulator. 

\end{itemize}

\section{Related Work}

\subsection{Mobile Manipulation Systems and Policies}

Mobile manipulation requires coordinating base motion and manipulation over long-horizon tasks. Existing approaches span modular systems that separate perception, navigation, and manipulation~\citep{yenamandra2023homerobot,liu2024okrobot}, and learned policies that couple mobility and manipulation through teleoperated demonstrations, scene representations, whole-body control, or trajectory models~\citep{fu2024mobilealoha,uppal2024spin,qiu2024geff,yan2024m2diffuser,sundaresan2025homer}.
With larger robot datasets and pretrained models~\citep{ye2026data}, generalist policies have scaled toward multi-task and cross-embodiment control~\citep{brohan2022rt1,openx2023rtx,ghosh2024octo,liu2024rdt1b,li2024robovlms}. Vision-language-action models further extend this paradigm to general robot control~\citep{brohan2023rt2,black2024pi0,black2025pi05,wen2025dexvla,gemini2025robotics,nvidia2025grootn1,nvidia2025grootn15,nvidia2025grootn16,zheng2025xvla,wu2026lingbotvla,yu2026walloss,xiaomi2026robotics1,yuan2026qwenrobotmanip,kim2026rldx1}, while recent work targets mobile manipulation through fixed-base VLA transfer, whole-body action prediction, spatially enriched observations, and action-conditioned world models~\citep{wu2025momanipvla,tu2026sgvla,yang2026panovla,kim2026cosmospolicy,yuan2026fastwam,chen2026abotm05,ma2026decowam}. System-level approaches additionally incorporate structured planning, manipulation-aware base positioning, and recovery~\citep{li2026momastage,lim2026anchorvla}. Our focus is complementary: we study how failures are distributed across task planning, navigation and base positioning, and local manipulation in an agentic system, and how the dominant bottlenecks can be addressed by agentic intervention. 

\subsection{Robot Failure Diagnosis and Recovery}

Task success rate alone provides limited information about why a robot fails. Recent benchmarks evaluate policies along fine-grained capability and failure dimensions~\citep{gao2026ebench,yu2026so101failure}, while diagnosis methods use multimodal execution traces or collected failure data to identify failure causes and generate corrective feedback~\citep{liu2023reflect,duan2025aha,lin2025failsafe,lu2025robofac,zeng2026vifailback}. Other methods integrate failure handling directly into execution through VLM supervisors, failure detectors, recovery policies, or world-model-based verification~\citep{chen2024failure_recovery_vlm,yang2026fpcvla,zhao2026flare,luo2026fdec,seligmann2026vlafail,liu2026checkvla,zhang2026robustexecution}. Rather than focusing on failure detection or recovery itself, we analyze where failures arise in an agentic mobile manipulation system, use this diagnosis to guide interventions, and track how the remaining failure distribution shifts as the system improves.

\subsection{Agentic Robots, Demonstrations, and Memory}

Language and vision-language models have been widely used for robot task planning, semantic grounding, and closed-loop execution~\citep{ichter2023saycan,huang2023innermonologue,wang2025karma,driess2023palme,huang2023voxposer,rana2023sayplan,chen2024autotamp,sun2026dadu}. Programmatic and coding-agent-based systems further compose perception, planning, and control tools through executable programs, iterative feedback, debugging, and skill refinement~\citep{liang2023code,singh2023progprompt,huang2023instruct2act,fu2026capx,lu2026aspire,xiao2026enpire,zhang2026harnessvla}, with recent work also exploring tighter integration between VLAs, external tools, and specialized robot policies~\citep{lei2026towards,shi2026memoryvla}.

Video demonstrations can provide task specifications or procedure planning for long-horizon manipulation~\citep{zhong2023procedure,wang2023mimicplay,jain2024vid2robot,wake2023gpt4vrobot}, while memory and reusable skills help agents retain task progress and reuse prior experience~\citep{yoneda2024statler,zhang2026harnessvla,shi2026memoryvla,sridhar2026memer,shah2026halo}. 
In our agentic framework, video-based demonstrations are leveraged to build a skill library for task planning, while visual memory is maintained for long-horizon tasks. They serve as the targeted interventions on top of a fixed subtask VLA.
\section{Failure Diagnosis of Agentic Mobile Manipulation}
\label{sec:diagnosis}

We analyze where the agentic baseline fails during long-horizon mobile manipulation. We first describe the evaluation tasks and baselines, then define the failure attribution procedure and summarize the main failure patterns.

\subsection{Diagnostic Setting}
\label{sec:diagnostic_setting}

\textbf{Benchmark and tasks.} We study all 20 composite tasks in RoboCasa365~\citep{nasiriany2026robocasa365} that require mobile manipulation. These tasks contain 2--15 subtasks and require the robot to move between workspaces while performing object transport, pick-and-place, and articulated-object interaction. Additional task details are provided in Appendix~\ref{app:task-set}.

\textbf{Baseline.} Our baseline follows a hierarchical agentic design adapted from Harness VLA~\citep{zhang2026harnessvla}. A high-level planner selects subtask instructions or tools, while a shared subtask VLA executes both navigation and manipulation. The subtask VLA is trained on RoboCasa365 atomic task demonstrations and subtask-level segments from the training split. Implementation and training details are provided in Appendix~\ref{app:vla_training}.

\textbf{Preliminary results.} While the subtask VLA can achieve 80\% success rate on the atomic tasks of RoboCasa365, the agentic system reaches only 19\% success rate on the mobile manipulation tasks of RoboCasa365. This gap motivates a key question: \emph{where does this agentic system fails?}

\subsection{Failure Attribution}
\label{sec:failure_attribution}

We evaluate the baseline on the 20 mobile manipulation tasks with five rollouts per task, giving 100 episodes in total. The baseline succeeds in 19 episodes, leaving 81 failed episodes for failure analysis. For each failed episode, we review the recorded actions, observations and simulator's state to identify failed subtasks. We then check whether the agent recovers from each error within the next three replanning attempts. If the affected subtask is completed within this window, we consider the error recovered and exclude it from primary failure attribution. Otherwise, we examine the remaining trajectory and identify the earliest error that remains unresolved and prevents task completion. 
We denote its time step as $t^*$. For example, if the robot fails to grasp an object and never picks it up, we attribute the failure to grasping rather than to a later placement attempt. If the grasp is recovered but placement later fails, we attribute the failure to placement.

For offline diagnosis, we examine the rollout video, planner interaction history, robot states, and simulator object states. Simulator states are used only for diagnosis and are not available to the online agent. Three independent Codex diagnoses provide candidate labels, which are checked by human reviewers using the recovery and attribution rules above. On a random subset of 30 failed episodes, three reviewers unanimously agree on 26 cases, with Fleiss' $\kappa=0.86$~\citep{fleiss1971measuring}. Further details and the annotation rubric are provided in Appendix~\ref{app:failure_attribution}.

\begin{figuretextblock}
\subsection{Failure Analysis}
\label{sec:failure_analysis}

\begin{figure}[H]
    \centering
    \includegraphics[width=0.9\linewidth]{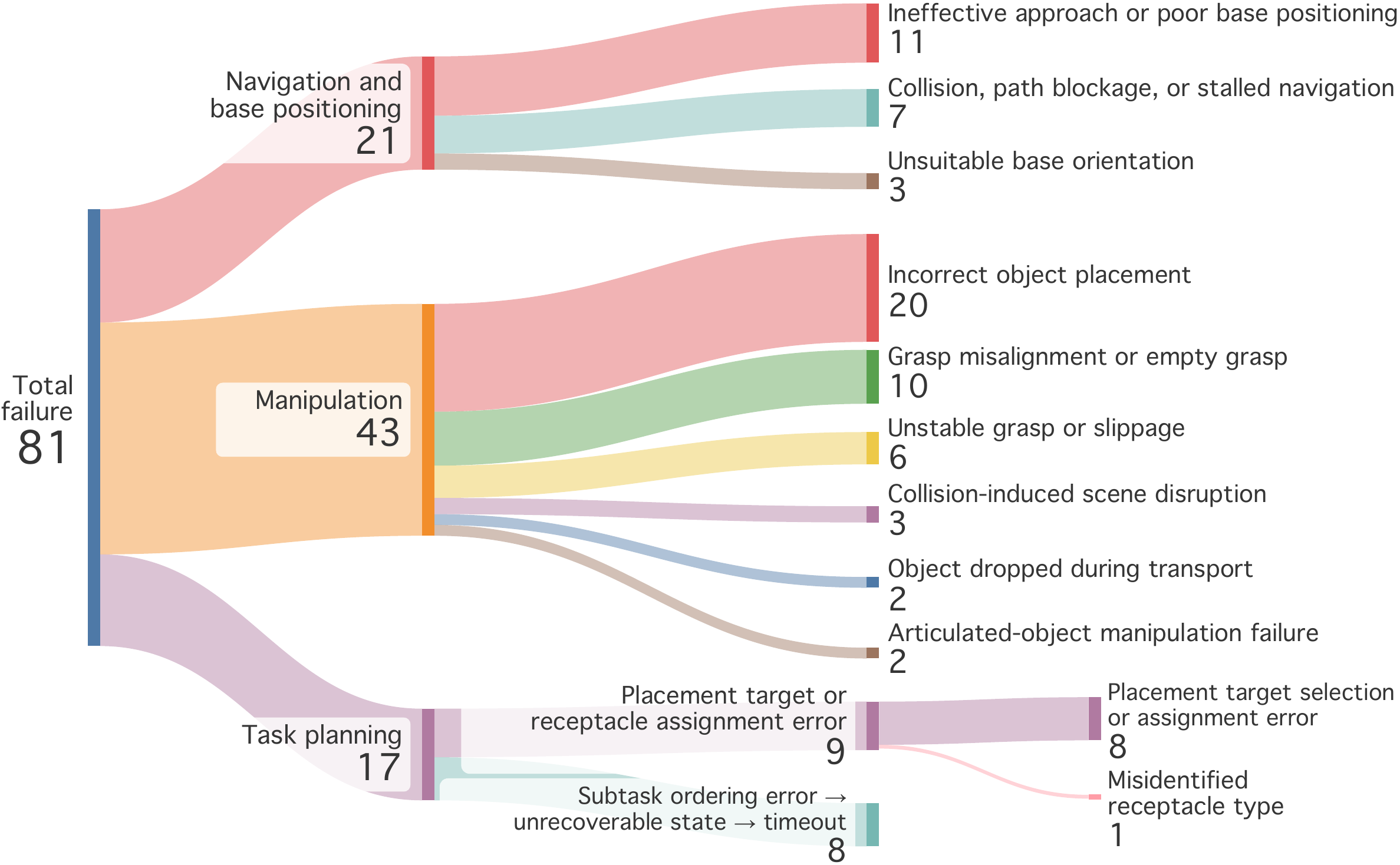}
    
    \caption{
    \textbf{Failure attribution of the agentic baseline.}
    Among the 81 failed episodes, the primary failures are attributed to task planning (21.0\%), navigation and base positioning (25.9\%), and local manipulation (53.1\%).
    }
    \label{fig:failure}
    
\end{figure}

\paragraph{Finding 1: Task planning fails when task progress is not tracked reliably.}
In many planning failures, the necessary low-level skills are already available, but the planner invokes them for the wrong target or at the wrong stage of the task. This problem is particularly pronounced in long-horizon execution, where an early action can alter the scene and invalidate subsequent decisions. Reliable task planning therefore requires maintaining an up-to-date representation of what has already been completed, what remains to be done, and how past actions have changed the scene~\citep{huang2023innermonologue,yoneda2024statler}.
\end{figuretextblock}

\paragraph{Finding 2: Reaching the correct workspace does not guarantee a suitable base pose for manipulation.}
A substantial fraction of failures occurs after the robot has navigated to the correct part of the scene, but stops at a base pose from which the next manipulation is difficult or even infeasible. Because the robot base determines the reachable workspace and viewing geometry of the arm, successful navigation should be judged not only by whether the robot reaches the target area, but also by whether it arrives at a pose that affords the upcoming interaction~\citep{yenamandra2023homerobot,liu2024okrobot}. This exposes a critical handoff problem between navigation and manipulation.

\paragraph{Finding 3: Precise manipulation can still fail after planning and navigation succeed.}
Even when the robot reaches a suitable workspace, the final interaction may fail because of target localization errors, grasp misalignment, or inaccurate end-effector positioning. These errors are often subtle in geometric terms but critical for the contact-rich actions such as grasping and placement. Therefore, successful mobile manipulation requires a final stage of local geometric adjustment that converts a semantically correct decision into a precise physical interaction~\citep{huang2023voxposer}.

\paragraph{Taken together,} these findings expose three weak links in the long-horizon mobile manipulation: {\textbf{tracking task progress}}, {\textbf{establishing a manipulation-ready base pose}}, and {\textbf{executing precise local interactions}}. Importantly, these failure modes are sequential in the task execution process: planning errors can prevent the robot from reaching the right workspace, poor base poses can prevent a correct subtask from being executed, and the small geometric errors can still derail the final interaction. This failure structure naturally motivates our failure-guided agentic interventions, which target each bottleneck at the stage where it arises.

\begin{figuretextblock}
\section{Failure-Guided Agentic Intervention}
\label{sec:method}

Guided by the above failure analysis, we keep the subtask VLA fixed and add failure-guided interventions at the bottlenecks. These interventions includes demonstration-based planning guidance, adaptive visual memory, an accessibility check with base adjustment, and local end-effector adjustment. Figure~\ref{fig:framework} summarizes the resulting failure-guided agentic framework.

\begin{figure}[H]
    \centering
    \includegraphics[width=0.99\linewidth]{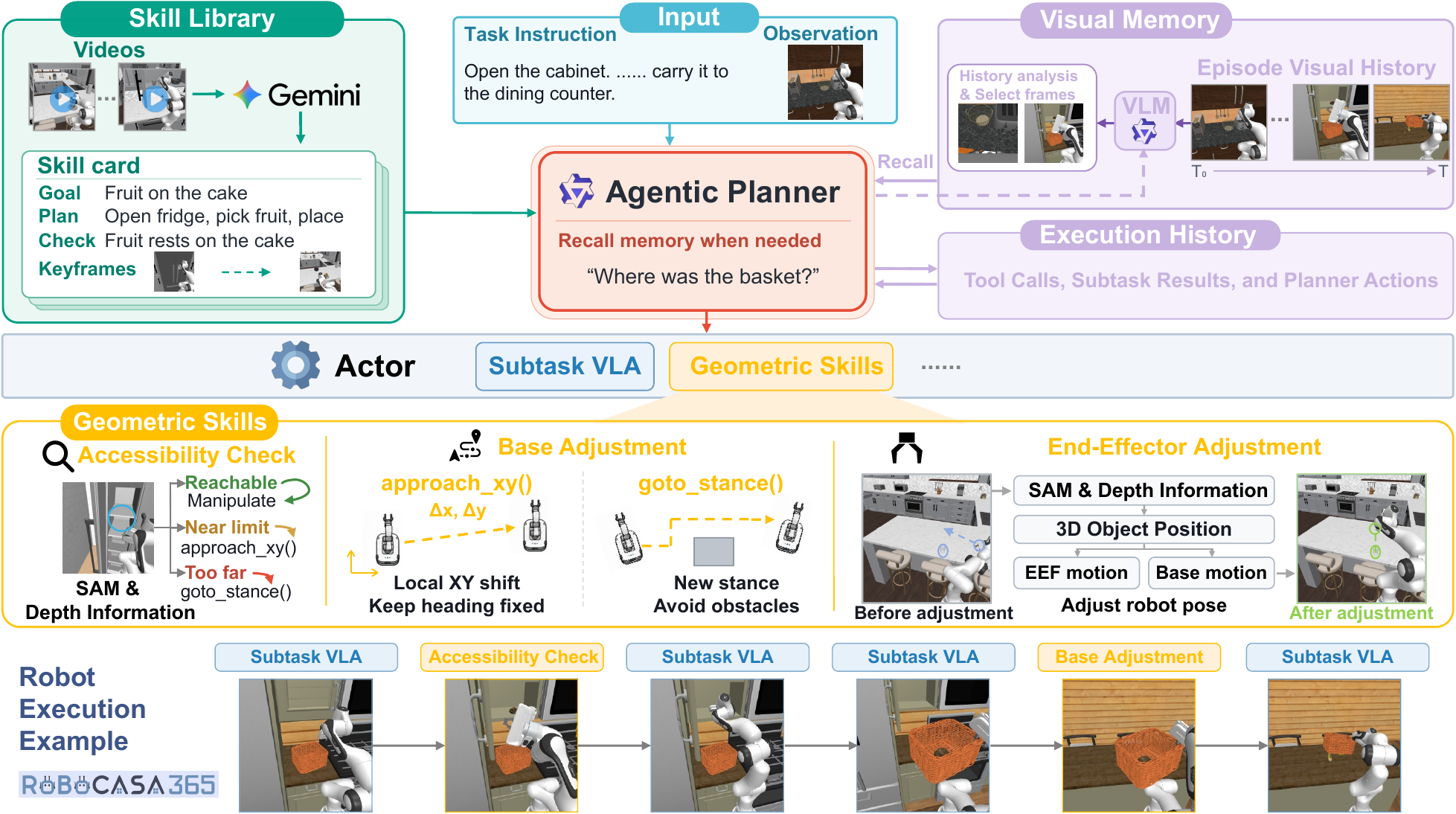}
    
    \caption{
    \textbf{Overview of the failure-guided agentic system.}
    The agentic planner leverages the skill library for task planning guidance, visual memory for relevant observation retrieval, and geometric skills for accessibility checking, base adjustment, and end-effector adjustment.
    }
    \label{fig:framework}
    
\end{figure}
\end{figuretextblock}

\subsection{Skill Library from Video Demonstrations}
\label{sec:demo}

Planning failures often leads to a wrong target or a wrong subtask order. We build a skill library from demonstration videos to guide the planner~\citep{zhong2023procedure,wake2023gpt4vrobot}. 
For each evaluation task, Gemini 3.5 Flash~\citep{google2026gemini35flash} matches task instruction to the annotations in the training data of corresponding benchmark and retrieves a demonstration. The model then summarizes it as a \emph{skill card} including task goal, subtask order, representative keyframes, and completion criteria. 

During execution, the planner conditions its next-step decision on both the skill card and the current execution context, including the current observation and action history. This provides explicit task-level guidance while preserving the planner's ability to adapt its actions to the current scene.

\subsection{Adaptive Visual Memory}
\label{sec:recall}

Textual tool histories do not always capture where objects have moved or whether an interaction succeeded~\citep{yoneda2024statler,sridhar2026memer,shah2026halo}. Our \emph{adaptive visual memory} stores observations from the current episode and retrieves them on demand.

When the planner requests past information, Qwen3.8-27B~\citep{qwen2026qwen38_27b} ranks the stored observations against the query and returns the 16 most relevant frames with a short trajectory summary. Only the past observations are eligible, while the memory has no access to future frames or simulator states. For example, the query ``Where was the basket last seen?'' retrieves earlier views of the basket and their associated actions.

\subsection{Geometric Skills}
\label{sec:geometry}

The geometric skills seek to address the problems in two common situations: the base stops at a pose unsuitable for manipulation, or the end-effector misses the target. We first check target accessibility from the current base pose. If needed, we adjust the base before applying a local end-effector correction~\citep{liu2024okrobot,li2026momastage}.

\paragraph{Accessibility check and base adjustment.}

Before transitioning from navigation to manipulation, the agent invokes \texttt{check\_stance} to determine whether the current base pose provides sufficient access to the target. The current subtask specifies the target object or interaction region, whose name is used as a text query for Grounding DINO~\citep{liu2024groundingdino}, followed by SAM~\citep{kirillov2023segment} to refine the resulting target mask. Valid depth pixels within the mask are back-projected using the calibrated camera parameters to estimate the target position in the robot coordinate frame. Based on the target distance, we classify the current stance as \emph{reachable}, \emph{marginal}, or \emph{out of reach}. The distance thresholds are estimated from successful training trajectories and fixed before evaluation.

While the accessibility check considers the target distance, it does not validate the joint limits, inverse kinematics, or full-body collisions. 
When the stance is \emph{marginal}, the agent invokes \texttt{approach\_xy} to apply a small local base correction while keeping the heading fixed. When the stance is \emph{out of reach}, it invokes \texttt{goto\_stance}, which localizes the target from the current observations, selects a new base stance within the manipulation range, and plans a path to that stance using $A^*$~\citep{hart1968formal} on a 2D occupancy grid constructed from the current depth observations. Before planning, observed obstacles are inflated by 0.1~m to provide a safety margin, and the robot follows the resulting waypoints using a PD controller. The distance thresholds and other geometric-skill parameters are fixed before evaluation and reported in Appendix~\ref{app:skill_details}.

\paragraph{End-effector adjustment.}

Even from a manipulation-ready base pose, small errors in target localization or geometric alignment can cause grasping and placement to fail. We thus provide local end-effector adjustment for correcting the final interaction pose. For each grasping or placement subtask, the target object or placement region is specified by the current subtask. 
We use the same Grounding DINO \& SAM pipeline to obtain the target mask, and back-project valid depth pixels within the mask into 3D. The median of these points is used as the target position, reducing sensitivity to depth outliers and missing depth measurements. We additionally estimate a local surface normal from the 3D points to determine the end-effector direction. 
The agent then applies a bounded correction toward an offset pose along this direction, with the translation magnitude limited to 0.3~m per call. 
The subtask VLA remains fixed, while the geometric skill only changes the local pose.

\section{Experiments and Results}
\label{sec:experiments}

We first study progressive integration and failure evolution on the development episodes of RoboCasa365. After fixing the system design, prompts, and tool parameters, we evaluate the finalized framework on held-out episode seeds with different scenes and object initializations. 
We then test the same agentic design on BEHAVIOR-1K~\citep{li2023behavior} and on  real-robot tasks. Finally, we compare against end-to-end and agentic baselines and examine key design choices through ablation study.

\subsection{Evaluation Protocol}
\label{sec:exp_setup}

We evaluate all methods under the same task setting in Section~\ref{sec:diagnosis}, using five rollouts per task. During development, all configurations are tested on the same episode seeds and simulator initial states, enabling direct paired comparison as components are added. 
After the framework design, prompts, and tool parameters are finalized, we conduct a separate evaluation using five held-out episode seeds per task. These held-out episodes use the same task definitions but different scene layouts and object initial states, providing a test of generalization beyond the development episodes.

Unless otherwise stated, all agentic methods use Kimi K3~\citep{team2026kimi} as the high-level agent and share the same fixed subtask VLA trained on RoboCasa365~\citep{nasiriany2026robocasa365}, so performance differences arise from the agentic design rather than the changes to the underlying manipulation policy. Task success rate is measured using the official RoboCasa365 evaluator. We additionally report paired statistical analyses in Appendix~\ref{app:stats}. To assess transfer beyond RoboCasa365, we further evaluate the same framework on BEHAVIOR-1K and on a real mobile manipulator.

\subsection{Progressive Integration}
\label{sec:progressive}

\begin{figure}[H]
    \centering
    
    \includegraphics[width=0.56\linewidth]{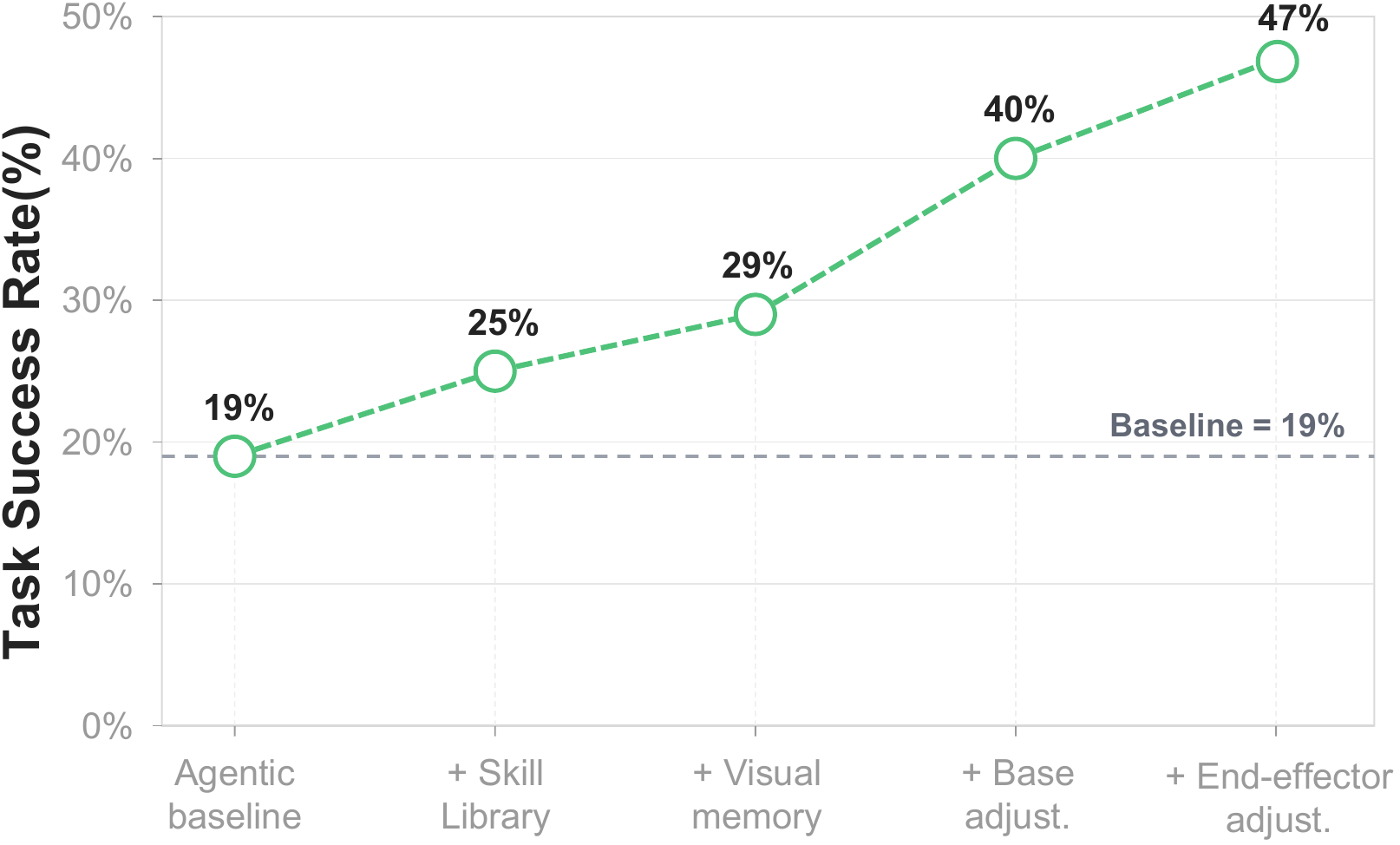}
    
    \caption{
    \textbf{Progressive integration of failure-guided interventions.}
    Each point adds one intervention group to the previous configuration.
    }
    \label{fig:progressive}
    
\end{figure}

Starting from the agentic baseline, we progressively introduce skill library, adaptive visual memory, base adjustment, and end-effector adjustment, while keeping the underlying subtask VLA fixed. 

As shown in Figure~\ref{fig:progressive}, this progressive integration raises task success from 19.0\% to 47.0\%. Base adjustment yields the largest incremental improvement in this sequence, increasing success from 29.0\% to 40.0\%.

\begin{figuretextblock}
\begin{figure}[H]
    \centering
    \includegraphics[width=0.62\linewidth]{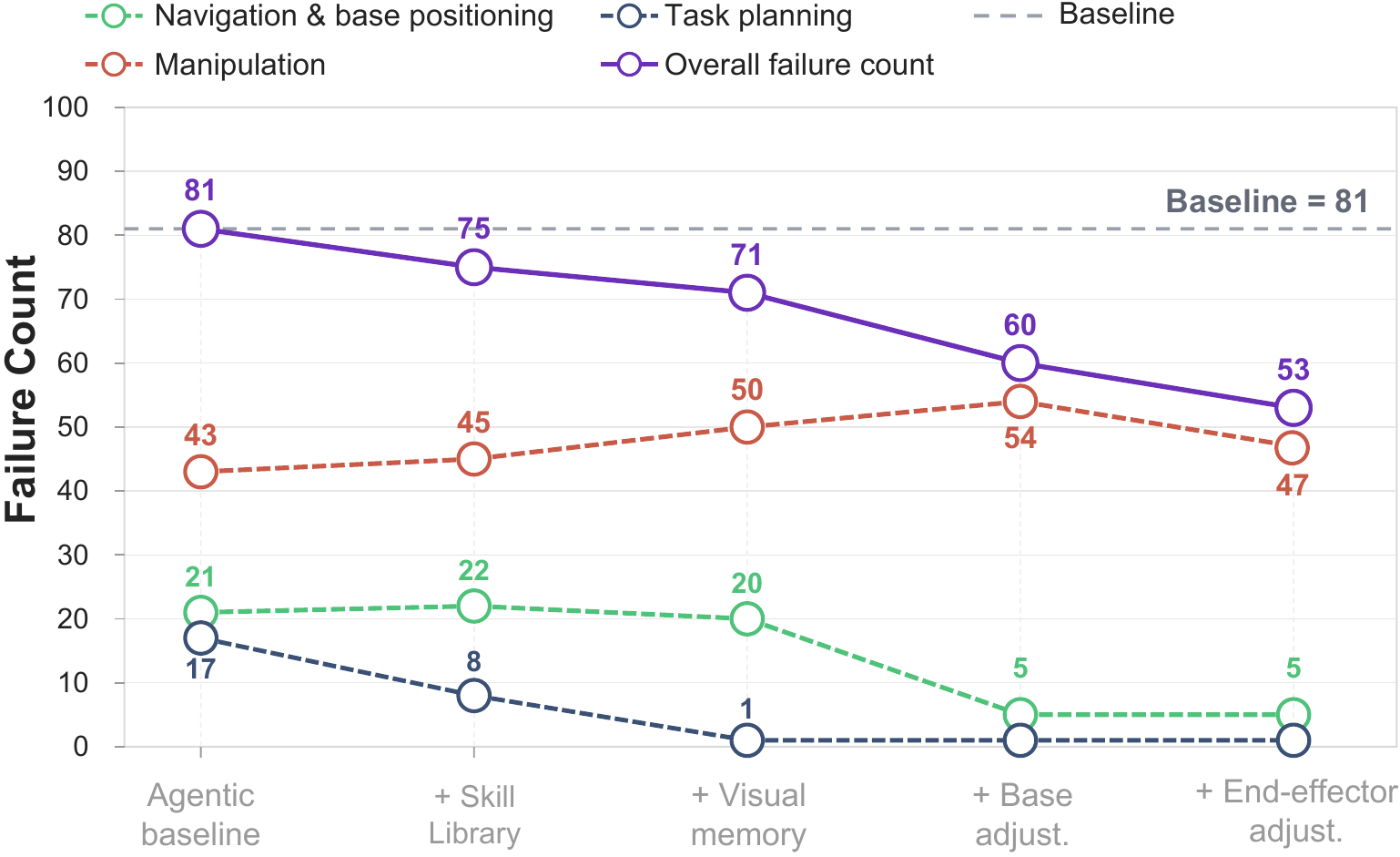}
    
    \caption{
    \textbf{Failure distribution during progressive integration.}
    Counts are reported over the same 100 matched episodes.
    }
    \label{fig:failure_attribution_progressive}
    
\end{figure}

Figure~\ref{fig:failure_attribution_progressive} shows the details of each failure type as these skills are added. The earlier planning-oriented interventions mainly reduce failures in target selection and task progression, whereas base adjustment substantially reduces navigation and base-positioning failures.
As planning and navigation bottlenecks are alleviated, more episodes reach later stages of execution, causing manipulation to emerge as the dominant remaining failure type. This shift provides direct evidence of a moving failure frontier: improving one stage exposes the next bottleneck along the pipeline.

\end{figuretextblock}

A fine-grained breakdown reveals the same trend. Failure modes explicitly targeted by our interventions are substantially reduced, including placement target or receptacle assignment errors (9$\rightarrow$0), ineffective approaches or poor base positioning (11$\rightarrow$1), and grasp misalignment or empty grasps (10$\rightarrow$3). 
The remaining failures are concentrated in cases such as incorrect object placement, object drops during transport, and collision-induced scene disruption. These errors occur at a later stage of execution and are not fully addressed by the current geometric corrections. The complete fine-grained failure attribution is provided in Appendix~\ref{app:full_failure_attribution}. Detailed task results and task horizon results can be found in Appendix~\ref{app:per_task}.

\begin{figuretextblock}
\subsection{Held-Out Evaluation}
\label{sec:main_comparison}

\begin{table}[H]
    \centering
    \caption{
        \textbf{Performance of mobile manipulation tasks on the RoboCasa365 benchmark.} We evaluate the 20 of 32 composite tasks that require mobile manipulation.
    }
    \label{tab:main_comparison}
    \small
    \setlength{\tabcolsep}{4.0pt}
    \begin{tabular}{llcc}
        \toprule
        Type & Method & Development (\%) & Held-out (\%) \\
        \midrule

        \multirow{6}{*}{VLA}
        
        & $\pi_{0}$~\citep{black2024pi0}
        & 2.0 & 2.0 \\
        
        & $\pi_{0.5}$~\citep{black2025pi05}
        & 3.0 & 4.0 \\

        & GR00T N1.6~\citep{nvidia2025grootn16}
        & 7.0 & 8.0 \\

        & GR00T N1.5~\citep{nvidia2025grootn15}
        & 8.0 & 9.0 \\

        & RLDX-1~\citep{kim2026rldx1}
        & 15.0 & 14.0 \\

        & Xiaomi Robotics 1~\citep{xiaomi2026robotics1}
        & 44.0 & 45.0 \\

        \midrule

        \multirow{4}{*}{Agentic}
        & CaP-X (Kimi K3)~\citep{fu2026capx}
        & 10.0 & 11.0 \\

        & Pigey (Kimi K3) ~\citep{galanti2026addressing}
        & 13.0 & 12.0 \\

        & Harness VLA (Kimi K3) ~\citep{zhang2026harnessvla}
        & 19.0 & 18.0 \\

        & \textbf{Ours (Kimi K3)}
        & \textbf{47.0} & \textbf{49.0} \\

        \bottomrule
    \end{tabular}
    
\end{table}

Table~\ref{tab:main_comparison} compares our full system against representative end-to-end VLA policies and agentic baselines. All methods are evaluated on the same episode seeds to ensure a controlled comparison. For agentic methods, we additionally match the planner budget and retry budget. Pigey and our Harness VLA adaptation use the same subtask VLA checkpoint as our method, isolating the effect of the agentic design. For end-to-end baselines, we use the released checkpoints and implementations linked from the official RoboCasa365 repository. Additional details are provided in Appendix~\ref{app:vla_training}.

\end{figuretextblock}

On the development episodes, our full system improves task success from 19.0\% to 47.0\% over the Harness VLA baseline. It is worth noting that this gain persists on the held-out evaluation: after fixing the system design, prompts, and tool parameters, success increases from 18.0\% to 49.0\% on unseen episode seeds with different scene and object initializations. The held-out improvement remains statistically significant under paired analysis ($p<0.001$) (Appendix~\ref{app:stats}), indicating that the gains are not specific to the episodes used during system development. Our full system also achieves performance comparable to the strongest end-to-end VLA baseline, Xiaomi Robotics 1, despite using a modular agentic design built around a fixed subtask VLA.

We further test whether the same failure-guided design transfers beyond RoboCasa365. On five BEHAVIOR-1K tasks, our framework improves task success from 14.0\% to 37.0\% over the corresponding Harness VLA baseline. For this evaluation, the low-level policies are trained on BEHAVIOR-1K training data, while the agentic intervention design remains unchanged. The experiment setup and detailed results of BEHAVIOR-1K can be found in Appendix~\ref{app:behavior1k}. 
On two real-robot tasks, the full system completes 13 of 20 trials, compared with 5 of 20 for the agentic baseline. Their task definitions and results are in Appendix~\ref{app:real_robot}.
Overall, these results suggest that the benefits of failure-guided agentic intervention are not tied to a single benchmark or scene distribution.

\subsection{Ablation Study}
\label{sec:ablation}

\begin{table}[!htbp]
    \centering
    \caption{
        \textbf{Design ablations on 20 mobile manipulation tasks from RoboCasa365.}
        Each group varies one design choice while keeping the remaining
        components of the full system unchanged.
    }
    \label{tab:agentic_ablation}
    \small
    \setlength{\tabcolsep}{4pt}

    \LayoutFit{%
    \begin{tabular}{llc}
        \toprule
        Component
        & Variant
        & Success Rate (\%) \\
        \midrule

        Reference
        & \textbf{Full system}
        & \textbf{47.0} \\

        \midrule
        \multirow{3}{*}{Skill Library}

        & Text-only skill card (VLA training data)
        & 43.0 \\
        
        & Text skill card + Keyframes (HIW-500)
        & 40.0 \\

        &
        Text skill card + keyframes (VLA training data)
        & \textbf{47.0} \\

        \midrule
        \multirow{6}{*}{Adaptive Visual Memory}
        & Uniform history every turn (8 frames)
        & 37.0 \\

        &
        Uniform history every turn (16 frames)
        & 40.0 \\

        &
        Uniform history every turn (32 frames)
        & 35.0 \\

        &
        Planner-triggered memory query (8 frames)
        & 42.0 \\

        &
        Planner-triggered memory query (16 frames)
        & \textbf{47.0} \\

        &
        Planner-triggered memory query (32 frames)
        & 45.0 \\

        \midrule
        \multirow{3}{*}{Base Adjustment}
        & w/o accessibility check (\texttt{check\_stance})
        & 36.0 \\

        &
        w/o local base positioning adjustment (\texttt{approach\_xy})
        & 42.0 \\

        &
        w/o stance replanning (\texttt{goto\_stance})
        & 39.0 \\

        \midrule
        \multirow{2}{*}{End-effector Adjustment}
        & Target location by agentic planner
        & 43.0 \\

        & Target location by geometric skills
        & \textbf{47.0} \\

        \bottomrule
    \end{tabular}%
    }
\end{table}

In this section, we examine the design choices. All ablations are evaluated on the same 100 matched episodes, and paired statistical comparisons are reported in Appendix~\ref{app:i_4}.

\begingroup
\titlespacing*{\paragraph}{0pt}{8pt plus 1pt}{0.6em}
\paragraph{Skill library.}

We study two design choices: the source of the demonstrations and whether visual keyframes are included in the skill card. 
HIW-500~\citep{bitrobot2026hiw500} provides real-world teleoperated household demonstrations collected with humanoid robots, whereas our in-domain demonstrations are drawn from the RoboCasa365 training split. When using the same text-based skill representation and demonstration keyframes, replacing HIW-500 demonstrations with in-domain demonstrations increases the observed success rate from 40.0\% to 47.0\%. With the same in-domain demonstrations, removing the keyframes and retaining only the textual skill card reduces success to 43.0\%. 
These results suggest that both domain alignment and visual context contribute to effective demonstration-based guidance.

\paragraph{Adaptive visual memory.}

We compare continuously exposing visual history to the planner against retrieving it only when needed. In the always-provided setting, 8, 16, or 32 uniformly sampled history frames are appended at every planner turn. In our adaptive design, the planner decides when memory is needed, formulates a query, and retrieves the specified number of frames most relevant to that query. Planner-triggered retrieval achieves a higher observed success rate across all frame budgets, while increasing the number of provided frames does not consistently improve performance. 
This suggests that the value of visual memory depends more on retrieving task-relevant evidence than on exposing the planner to a larger amount of history.

\paragraph{Base adjustment.}

We ablate the main components of the base-adjustment pipeline to understand which stages are most important at the navigation--manipulation interface. Removing the accessibility check produces the largest degradation among the variants, while removing stance replanning has a larger effect than removing the local base-position correction. This pattern is consistent with our failure analysis: the key challenge is first to recognize when the current stance is unsuitable for manipulation and, when necessary, to reposition the robot to a new manipulation-ready stance.

\paragraph{End-effector adjustment.}

Finally, we compare two ways of specifying 3D target for local end-effector correction. Allowing the high-level planner to directly specify the adjustment target yields 43.0\% success rate, compared with 47.0\% when the target is estimated using our RGB-D geometric grounding pipeline. Qualitative rollouts are shown in Appendix~\ref{app:qualitative}. 
Although the numerical difference is modest, the result suggests that explicit geometric grounding can reduce the burden on the planner for precise spatial specification, providing a more reliable interface for local manipulation correction.

\endgroup

\section{Conclusion}
\label{sec:conclusion}

We presented a failure-guided agentic approach to long-horizon mobile manipulation that moves beyond simply improving the underlying manipulation policy. 
Starting from a pretrained VLA with strong subtask-level performance, we systematically identified where mobile manipulation breaks down and found three dominant failure regimes: task planning, navigation and base positioning, and local manipulation. 
We then translated these observations into agentic interventions, including demonstration-based planning guidance, adaptive visual memory, accessibility-aware base adjustment, and local end-effector adjustment. Across RoboCasa365, BEHAVIOR-1K, and real-robot experiments, these interventions substantially improve long-horizon task success rate while leaving the underlying subtask VLA fixed. 
This provides a useful perspective on agentic robot learning: capability progress can come not only from making the policy stronger, but from understanding what currently limits it and intervening at that bottleneck.

More broadly, we hope this work encourages failure diagnosis to become a first-class tool for designing agentic embodied systems. As robots are deployed in increasingly long-horizon, multi-stage environments, their dominant failure modes may evolve throughout execution and as capabilities improve. Identifying and addressing these changing bottlenecks could provide a systematic path toward more reliable and capable mobile manipulation.
\FloatBarrier
\section*{AI Use Statement}

Generative AI tools were used to assist with literature collection, partial manuscript drafting and language polishing, as well as the implementation and debugging of parts of the experimental code. The research ideas, methodology, experimental design, analysis, and conclusions were developed and verified by the authors. All experimental results reported in this paper were obtained from the authors' actual experiments.

\section*{Reproducibility Statement}
The appendix reports the evaluation task set, subtask VLA training setup, failure attribution procedure, and additional agent-side skill details. Code, prompts, configuration files, and evaluation episode lists will be released to support reproduction of the reported experiments.

\FloatBarrier
\bibliographystyle{assets/plainnat}
\bibliography{paper}

\FloatBarrier
\pagebreak[4]
\beginappendix
\begingroup
\small

\setlength{\parskip}{4pt plus 1pt}
\setlength{\intextsep}{8pt plus 1pt minus 1pt}
\captionsetup{skip=4pt}
\renewcommand{\arraystretch}{1.08}
\FloatBarrier
\section{Mobile Manipulation Evaluation Tasks}
\label{app:task-set}

We evaluate all 20 RoboCasa365~\citep{nasiriany2026robocasa365} benchmark tasks that require mobile manipulation. The tasks cover navigation between workspaces, object pickup and placement, object transport, interaction with articulated objects, and manipulation around appliances and receptacles. Following the official RoboCasa365 decomposition, the tasks contain 2--15 subtasks, with 91 subtasks in total and an average of 4.55 subtasks per task. The task names, subtask counts, and per-task results are reported together in Appendix~\ref{app:per_task} to avoid a separate task-list table.

\FloatBarrier
\section{Agentic Baseline and Subtask VLA Details}
\label{app:vla_training}

\subsection{Agentic Baseline}
\label{app:agentic_baseline}

We adapt the tool-based agentic system of Harness VLA~\citep{zhang2026harnessvla}. Kimi K3~\citep{team2026kimi} serves as the high-level planner in a closed perception--reasoning--action loop. At each planner step, it either issues a short subtask instruction to the VLA or invokes an available execution tool. Each VLA call executes for 350 control steps before returning control to the planner. The planner then observes the updated scene and decides whether to continue the current subtask, retry it, or proceed to the next one. Navigation and manipulation use the same subtask VLA checkpoint and differ only in the instruction issued by the planner. We retain the planning and closed-loop execution structure of Harness VLA, while removing simulator-specific tools that are not readily transferable to real robots. The same planner interface and execution settings are used across all agentic configurations unless otherwise stated.

\subsection{Subtask VLA Training}
\label{app:subtask_vla}

\paragraph{Training data.}
We train a single subtask-conditioned VLA using only the official RoboCasa365~\citep{nasiriany2026robocasa365} training split. The training data combine atomic-task demonstrations with subtask level segments extracted from composite task demonstrations. The composite portion contains 16,181 demonstrations and 12.73M frames at 20~Hz. After removing the terminal \texttt{done} stage, 12.27M frames remain, yielding 84,401 valid subtask segments. No evaluation trajectory is used for policy training.

\begin{table}[H]
    \centering
    \caption{
        \textbf{Training settings of the subtask VLA.} The same checkpoint is used by all agentic configurations unless otherwise stated.
    }
    \label{tab:subtask_vla_training}
    \small
    \setlength{\tabcolsep}{5pt}
    \begin{tabular}{lc}
        \toprule
        Setting & Value \\
        \midrule

        VLM backbone
        & Qwen3-VL-4B-Instruct \\

        Visual encoder
        & DINOv2 ViT-S/14 \\

        Action model
        & DiT-B \\

        Image resolution
        & $224\times224$ \\

        State dimension
        & 16 \\

        Action dimension
        & 12 \\

        Action horizon
        & 50 \\

        DiT hidden size
        & 1024 \\

        DiT layers
        & 16 \\

        Inference steps
        & 10 \\

        Batch size
        & 256 \\

        Training steps
        & 200,000 \\

        Warmup steps
        & 1,000 \\

        VLM learning rate
        & $2.5\times10^{-5}$ \\

        Action-model learning rate
        & $1.0\times10^{-4}$ \\

        LR scheduler
        & Cosine \\

        Minimum learning rate
        & $2.5\times10^{-6}$ \\

        Optimizer
        & AdamW~\citep{loshchilov2019decoupled} \\

        AdamW $\beta$
        & $(0.9,\,0.95)$ \\

        Gradient clipping
        & 1.0 \\

        Random seed
        & 42 \\

        \bottomrule
    \end{tabular}
\end{table}

\paragraph{Subtask segmentation.}
Each frame in a composite demonstration is annotated with a subtask index, a natural-language subtask instruction, an atomic skill label, and an execution stage. We split each trajectory at the annotated subtask boundaries and condition each segment on its corresponding subtask instruction. Action chunks are truncated at segment boundaries and padded with the final action when needed, so a training sample never crosses into the next subtask. The resulting dataset contains 734 distinct subtask instructions.

\paragraph{Model and training.}
We implement the subtask VLA with StarVLA~\citep{starvla2026} using the QwenPI architecture. The vision-language backbone is Qwen3-VL-4B-Instruct~\citep{bai2025qwen3vl}, together with a DINOv2 ViT-S/14 visual encoder~\citep{oquab2024dinov2} and a DiT-B action head~\citep{peebles2023scalable}. Each training sample contains three RGB views resized to $224\times224$, the robot state, a subtask instruction, and a 50-step action chunk. The policy predicts the 12-dimensional RoboCasa action space from a 16-dimensional robot state. The action head uses a hidden size of 1024, 16 DiT layers, and 10 inference steps.

Atomic-task and subtask-level samples are jointly trained in a single checkpoint. We balance the two data sources so that frames in the combined dataset have the same long-run sampling probability, and sample without replacement within each epoch. Navigation and manipulation use the same checkpoint and differ only in the subtask instruction. The trained model achieves 80\% success on the RoboCasa365 atomic-task evaluation.

\paragraph{End-to-end baseline checkpoints.}
For the end-to-end baselines in Table~1, we use the released checkpoints and corresponding implementations linked from the official RoboCasa365 leaderboard. Table~\ref{tab:baseline_training_settings} summarizes the training settings reported by the corresponding submissions. We evaluate these released checkpoints on the same 20 mobile manipulation tasks and matched episode seeds used in our evaluation.

\begin{table}[H]
    \centering
    \caption{
        \textbf{Training settings of the released end-to-end baselines.}
        Batch sizes and training steps follow the corresponding
        RoboCasa365 leaderboard submissions.
    }
    \label{tab:baseline_training_settings}
    \small
    \setlength{\tabcolsep}{5pt}
    \begin{tabular}{lcc}
        \toprule
        Method
        & Batch Size
        & Training Steps \\
        \midrule

        $\pi_0$
        & 64
        & 75,000 \\

        $\pi_{0.5}$
        & 64
        & 75,000 \\

        GR00T N1.5
        & 128
        & 120,000 \\

        GR00T N1.6
        & 128
        & 120,000 \\

        RLDX-1
        & 192
        & 250,000 \\

        Xiaomi Robotics 1
        & 512
        & 120,000 \\

        \bottomrule
    \end{tabular}
\end{table}

\FloatBarrier
\section{Failure Attribution Details}
\label{app:failure_attribution}

\subsection{Attribution Procedure and Annotation Consistency}
\label{app:annotation_consistency}

\paragraph{Reviewing failed rollouts.}
For each failed episode, we inspect the rollout video, planner history, robot states, and simulator object states. Simulator states are used only for offline diagnosis. We compare the expected result of each subtask with the recorded state after execution. A failed grasp, missed placement, or unsuitable base position is marked as a candidate error. We also check whether the planner selects the wrong target or skips a required subtask.

\paragraph{Checking recovery.}
We examine the next three replanning attempts after each candidate error. Each attempt consists of a new high-level decision followed by an executed action and a new observation. Low-level control steps within a VLA call and read-only tool queries are not counted as separate attempts. An error is recovered if a retry or an alternative action completes the affected subtask within this window. If the episode ends sooner, we use the available attempts and the final state. For errors not recovered within three attempts, we inspect the remaining trajectory and exclude any error that is later corrected.

\paragraph{Selecting the primary failure.}
We check which task conditions remain unmet at the end of the episode. Among the errors that remain unresolved and prevent the task from being completed, we select the earliest one as the primary failure. We assign the label to the original error rather than to a later consequence. For example, if an unsuccessful grasp leaves the gripper empty and the object is never picked up, a subsequent failed placement is attributed to grasping. If the robot later grasps the object successfully but fails to place it, we instead attribute the failure to placement. Likewise, if an unsuitable base position is never corrected, later grasp attempts that fail because the object is inaccessible are attributed to base positioning.

\paragraph{Model-assisted review.}
We run Codex (GPT-5.6 Sol) three times on each failed rollout using the same diagnosis instructions. Three human reviewers check the suggested failure points against the recorded states and select one primary failure category per episode. These three offline diagnoses are separate from the three replanning attempts used to check recovery. For each episode, the annotation records the failed subtask, the expected and observed states, any recovery, the primary failure point, and its coarse and fine-grained labels.

\paragraph{Primary failure attribution consistency.}
To examine the consistency of the primary failure attribution, we randomly sample 30 failed episodes from the agentic baseline. Three reviewers independently assign each episode to one of the three primary failure categories used in Section~\ref{sec:failure_attribution}: task planning, navigation and base positioning, or manipulation.

The reviewers agree on 26 of the 30 episodes (86.7\%), with a Fleiss' $\kappa$ of 0.86. The remaining four cases are resolved by joint review before obtaining the final labels. Table~\ref{tab:failure_agreement} summarizes the annotation results.

\begin{table}[H]
    \centering
    \caption{
        \textbf{Agreement of independent failure attribution on a random subset of baseline failures.}
    }
    \label{tab:failure_agreement}
    \small
    \setlength{\tabcolsep}{5pt}
    \begin{tabular}{lcc}
        \toprule
        Failure Category & Episodes & Unanimous \\
        \midrule
        Task planning & 7 & 6 \\
        Navigation and base positioning & 8 & 7 \\
        Manipulation & 15 & 13 \\
        \midrule
        Total & 30 & 26 \\
        \bottomrule
    \end{tabular}
\end{table}

\paragraph{Fine-grained attribution consistency.}
We further evaluate the consistency of the fine-grained failure labels used in Figure~\ref{fig:failure}. We randomly sample 30 failed episodes across the evaluated agentic configurations. For this evaluation, the system configuration and enabled skills are hidden from the reviewers.

Three reviewers independently assign each episode to one of the fine-grained failure categories based on the rollout video, agent history, robot states, and simulator object states. The reviewers unanimously agree on 24 of the 30 episodes (80.0\%), with a Fleiss' $\kappa$ of 0.79~\citep{fleiss1971measuring}. The blinded majority label also agrees with the original fine-grained attribution on 27 of the 30 episodes (90.0\%).

The remaining disagreements mainly occur between closely related execution failures, such as incorrect placement and grasp or transport failures. Overall, the fine-grained labels remain consistent when the system configuration is hidden from the reviewers.

\begin{table}[H]
    \centering
    \caption{
        \textbf{Agreement of fine-grained failure attribution.} Three reviewers independently annotate 30 sampled failed episodes without access to the system configuration or enabled skills.
    }
    \label{tab:fine_grained_agreement}
    \small
    \setlength{\tabcolsep}{5pt}
    \begin{tabular}{lcc}
        \toprule
        Fine-grained failure category & Episodes & Unanimous \\
        \midrule
        Placement target or receptacle assignment & 4 & 3 \\
        Subtask ordering & 3 & 2 \\
        Ineffective approach or poor base positioning & 4 & 4 \\
        Collision, path blockage, or stalled navigation & 3 & 2 \\
        Unsuitable base orientation & 2 & 2 \\
        Incorrect object placement & 5 & 4 \\
        Grasp misalignment or empty grasp & 3 & 3 \\
        Unstable grasp or slippage & 2 & 1 \\
        Collision-induced scene disruption & 2 & 1 \\
        Object dropped during transport & 1 & 1 \\
        Articulated-object manipulation failure & 1 & 1 \\
        \midrule
        Total & 30 & 24 \\
        \bottomrule
    \end{tabular}
\end{table}

\FloatBarrier
\Needspace*{5\baselineskip}
\section{Skill Implementation Details}
\label{app:skill_details}

This appendix provides additional implementation details for the skill library, adaptive visual memory, and geometric skills used in the main experiments.

\subsection{Demonstration Retrieval and Skill Cards}

For each evaluation task, we retrieve a demonstration from the RoboCasa365 benchmark training split. The retrieval query uses the task instruction, and the candidate demonstrations are compared using their task annotations. After retrieval, Gemini 3.5 Flash~\citep{google2026gemini35flash} summarizes the demonstration into a skill card containing the task goal, ordered subtask plan, representative keyframes, and completion criteria. Skill cards are generated once before evaluation and remain fixed during rollout. The planner receives the skill card together with the current observation and execution history when planning the next step. No evaluation trajectory, evaluation outcome, or simulator state is used for demonstration retrieval or skill-card generation.

\subsection{Adaptive Visual Memory}

When the planner calls \texttt{recall\_memory(q)}, the recall model receives the natural-language query together with the stored candidate frames. It ranks the frames according to their relevance to the query and returns the top-$k$ frames together with a short summary of the corresponding trajectory history. We use $k=16$ in the full system unless otherwise stated in the ablation study.

The recall model uses Qwen3.8-27B~\citep{qwen2026qwen38_27b}, and candidate frames are resized to $448\times448$ before being passed to the model. Recall only operates on observations collected before the current agent step. We keep at most 256 candidate frames for each episode. When more than 256 observations are available, we uniformly subsample the past observations to form the candidate set. The recall model ranks these candidates conditioned on the agent query, and the top-k frames are returned to the planner. Adaptive recall means that the planner decides when to call the recall tool and provides a query based on its current information need; fixed recall instead provides uniformly sampled history frames without an agent-generated query.

\subsection{Geometric Skill Parameters}

The geometric skills use the RGB-D observations available to the online agent. Grounding DINO~\citep{liu2024groundingdino} first localizes the object or interaction region specified by the current subtask, and SAM~\citep{kirillov2023segment} refines the corresponding mask. Valid depth pixels are then back-projected into the robot coordinate frame.

For the accessibility check, a target within 0.75\,m of the robot is classified as reachable. Targets between 0.75\,m and 1.00\,m are classified as marginal, while targets beyond 1.00\,m are treated as out of reach. For a marginal stance, \texttt{approach\_xy} applies a local base correction of at most 0.25\,m while keeping the base heading fixed. The 0.75 m and 1.00 m accessibility thresholds are estimated from successful trajectories in the training split and are fixed before evaluation.

For an out-of-reach stance, \texttt{goto\_stance} constructs a 2D occupancy grid from the current depth observation and searches for a new base pose within the manipulation range. The occupancy grid uses a resolution of 0.05\,m. Obstacles are inflated by 0.10\,m before A* planning. 

For end-effector adjustment, the median 3D position of the grounded region is used as the target position. The local surface normal is estimated from the same point set and used to define the approach direction. Each adjustment is bounded to a maximum translation of 0.30\,m. Table~\ref{tab:skill_impl} summarizes the main settings used by the adaptive visual memory and geometric skills.

\begin{table}[!htbp]
    \centering
    \caption{
        Implementation settings for adaptive visual memory and geometric skills.
    }
    \label{tab:skill_impl}
    \small
    \setlength{\tabcolsep}{5pt}
    \begin{tabular}{ll}
        \toprule
        Parameter & Value \\
        \midrule

        \multicolumn{2}{l}{\textit{Adaptive visual memory}} \\

        Maximum candidate frames
        & 256 \\

        Retrieved frames ($k$)
        & 16 \\

        Recall model
        & Qwen3.8-27B \\

        Input image resolution
        & $448\times448$ \\

        \midrule
        \multicolumn{2}{l}{\textit{Geometric skills}} \\

        Reachable distance & $d\leq 0.75$\,m \\

        Marginal distance & $0.75<d\leq 1.00$\,m \\

        Maximum local XY shift
        & 0.25\,m \\

        Obstacle inflation
        & 0.10\,m \\

        Maximum EEF correction
        & 0.30\,m \\

        Occupancy-grid resolution
        & 0.05\,m \\

        \bottomrule
    \end{tabular}
\end{table}

\subsection{Planner Prompt Excerpts}
\label{app:planner_prompt}

The following summary covers the planner instructions for task execution, demonstration use, adaptive visual memory, and local correction. 

\begin{quote}
\small

\textbf{Task execution.}
Read the task instruction and current observations. Issue one short subtask instruction per VLA call. After execution, inspect the returned observations before choosing the next subtask. A completed tool call does not by itself mean that the subtask is complete.

\medskip
\textbf{Demonstration guidance.}
When a skill card is provided, use its plan and keyframes as a reference. Compare the current observations with the expected outcome before continuing. The demonstration scene may differ from the current scene; use the demonstrated behavior as guidance rather than copying object coordinates.

\medskip
\textbf{Adaptive visual memory.}
Query memory when past observations are needed to understand task progress or a failed attempt. Retrieved frames describe earlier states; use the latest observations to determine the current scene state.

\medskip
\textbf{Failure recovery.}
After an empty grasp or an off-target placement, inspect what happened before retrying. A closed gripper alone does not confirm a successful grasp. Localize the target from the current RGB-D observations rather than using fixed coordinates. Use local end-effector correction to address misalignment, then retry with the subtask VLA.

\medskip
\textbf{Tool calls.}
Briefly state the relevant observation and the chosen action before a tool call. Do not request another observation when the preceding tool has already returned an updated one.

\end{quote}

\FloatBarrier
\section{Paired Statistical Analysis}
\label{app:stats}

All progressive integration, leave-one-out, and fine-grained ablation experiments use matched episode seeds and simulator initial states across configurations. We therefore use episode-level paired analyses rather than treating success rates from two configurations as independent binomial samples.

For configurations $A$ and $B$, let $n_{01}$ be the number of matched episodes that fail under $A$ and succeed under $B$, and let $n_{10}$ be the number that succeed under $A$ and fail under $B$. The paired change in task success rate is
\[
\Delta_{\mathrm{succ}} = \frac{n_{01}-n_{10}}{N},
\]
where $N=100$ for the RoboCasa365 evaluations. We use a two-sided exact McNemar test~\citep{mcnemar1947note} for paired binary outcomes. Confidence intervals for the paired success-rate difference are estimated with paired bootstrap resampling~\citep{efron1979bootstrap} stratified by task, preserving the 20 task, five seed structure.

\subsection{Progressive Integration}

\begin{table}[H]
    \centering
    \caption{\textbf{Paired analysis of progressive integration.}
    The first four rows compare consecutive configurations; the final row compares the baseline with the full system on the same development episodes.
    $\Delta$ is reported in percentage points.}
    \label{tab:paired_progressive}
    \small
    \begin{tabular}{lcccccc}
        \toprule
        Comparison & Success (\%) & $\Delta$ (pp) & $n_{01}$ & $n_{10}$ & 95\% CI (pp) & McNemar $p$ \\
        \midrule
        Baseline $\rightarrow$ +Demo      & 19 $\rightarrow$ 25 & +6  & 8  & 2 & [0, 12]   & 0.109 \\
        +Demo $\rightarrow$ +Memory       & 25 $\rightarrow$ 29 & +4  & 7  & 3 & [-2, 10]  & 0.344 \\
        +Memory $\rightarrow$ +Base       & 29 $\rightarrow$ 40 & +11 & 15 & 4 & [3, 19]   & 0.019 \\
        +Base $\rightarrow$ +EEF          & 40 $\rightarrow$ 47 & +7  & 10 & 3 & [0, 14]   & 0.092 \\
        Baseline $\rightarrow$ Full       & 19 $\rightarrow$ 47 & +28 & 30 & 2 & [18, 38]  & $<0.001$ \\
        \bottomrule
    \end{tabular}
\end{table}

The baseline-to-full comparison shows a large paired improvement of 28 percentage points. Among the individual integration steps, base adjustment has the clearest paired effect. The smaller changes from skill library, adaptive visual memory, and end-effector adjustment have wider confidence intervals, so we treat them as supporting component evidence rather than separate strong statistical claims.

\subsection{Leave-One-Out and Held-Out Evaluation}

\begin{table}[H]
    \centering
    \caption{\textbf{Paired leave-one-out analysis on development and held-out episodes.}
    ``Memory'' denotes adaptive visual memory.}
    \label{tab:paired_loo}
    \small
    \LayoutFit{
    \begin{tabular}{llcccccc}
        \toprule
        Split & Comparison & Success (\%) & $\Delta$ (pp) & $n_{01}$ & $n_{10}$ & 95\% CI (pp) & McNemar $p$ \\
        \midrule
        Development
        & $-$Demo $\rightarrow$ Full    & 39 $\rightarrow$ 47 & +8  & 11 & 3 & [1, 15]  & 0.057 \\
        & $-$Memory $\rightarrow$ Full  & 41 $\rightarrow$ 47 & +6  & 9  & 3 & [-1, 13] & 0.146 \\
        & $-$Base $\rightarrow$ Full    & 35 $\rightarrow$ 47 & +12 & 14 & 2 & [5, 20]  & 0.004 \\
        & $-$EEF $\rightarrow$ Full     & 40 $\rightarrow$ 47 & +7  & 10 & 3 & [0, 14]  & 0.092 \\
        \midrule
        Held-out
        & Baseline $\rightarrow$ Full   & 18 $\rightarrow$ 49 & +31 & 33 & 2 & [21, 41] & $<0.001$ \\
        & $-$Demo $\rightarrow$ Full    & 40 $\rightarrow$ 49 & +9  & 12 & 3 & [2, 16]  & 0.035 \\
        & $-$Memory $\rightarrow$ Full  & 44 $\rightarrow$ 49 & +5  & 9  & 4 & [-2, 12] & 0.267 \\
        & $-$Base $\rightarrow$ Full    & 38 $\rightarrow$ 49 & +11 & 14 & 3 & [3, 19]  & 0.013 \\
        & $-$EEF $\rightarrow$ Full     & 42 $\rightarrow$ 49 & +7  & 11 & 4 & [0, 14]  & 0.118 \\
        \bottomrule
    \end{tabular}}
\end{table}

The leave-one-out results show the same overall pattern on development and held-out episodes.
Base adjustment has the largest leave-one-out effect in both splits. Most importantly, after the
system design, prompts, retrieval settings, and tool parameters are fixed, the full system improves
from 18.0\% to 49.0\% over the agentic baseline on held-out episodes, with a paired difference of
31 percentage points and $p<0.001$.

\subsection{Failure Category of Recovered Episodes}

To check whether each intervention mainly addresses the failure category that motivated it, we
inspect matched episodes that change from failure to success when the intervention is added.
Table~\ref{tab:recovered_failures} reports the primary failure category before the intervention.

\begin{table}[H]
    \centering
    \caption{\textbf{Primary failure types of episodes recovered after each intervention.}}
    \label{tab:recovered_failures}
    \small
    \begin{tabular}{lcccc}
        \toprule
        Added intervention & Recovered & Planning & Nav./Base & Manip. \\
        \midrule
        Skill library              & 8  & 5 & 1  & 2 \\
        Adaptive visual memory     & 7  & 4 & 1  & 2 \\
        Base adjustment            & 15 & 0 & 11 & 4 \\
        End-effector adjustment    & 10 & 0 & 1  & 9 \\
        \bottomrule
    \end{tabular}
\end{table}

The recovered episodes largely match the intended failure-to-intervention mapping: skill library and adaptive visual memory mainly recover task-planning failures, base adjustment mainly recovers navigation and base-positioning failures, and end-effector adjustment mainly recovers manipulation failures. An intervention can still affect later stages of a long-horizon episode, so this analysis should be interpreted at the episode level rather than as a one-to-one causal decomposition.

\FloatBarrier
\begingroup
\section{Paired Failure Transitions}
\label{app:failure_transition}

To complement the aggregate failure counts in Figure \ref{fig:failure_attribution_progressive}, we track the outcome of each matched episode from the baseline to the full system. Table~\ref{tab:failure_transition} shows transitions among success, task-planning failure, navigation/base-positioning failure, and manipulation failure.

\begin{table}[!htbp]
\centering
\caption{
Paired episode-level outcome transitions from the agentic baseline
to the full system.
}
\label{tab:failure_transition}
\small
\setlength{\tabcolsep}{5pt}
\begin{tabular}{lccccc}
\toprule
Baseline $\rightarrow$ Full
& Success
& Planning
& Nav./Base
& Manip.
& Total \\
\midrule
Success
& 17 & 0 & 1 & 1 & 19 \\
Planning
& 9 & 1 & 1 & 6 & 17 \\
Nav./Base
& 12 & 0 & 3 & 6 & 21 \\
Manipulation
& 9 & 0 & 0 & 34 & 43 \\
\midrule
Total
& 47 & 1 & 5 & 47 & 100 \\
\bottomrule
\end{tabular}
\end{table}


The transition table makes this survivor effect explicit. Six of the 17 baseline planning failures and six of the 21 navigation/base-positioning failures become manipulation failures in the full system. These transitions show that part of the increase in manipulation failures comes from episodes reaching later stages after
earlier failures are addressed. They do not determine whether the
underlying manipulation failure rate also changed. The transition counts also separate gains from regressions. The full system recovers 30 of the 81 baseline failures (37.0\%) and retains 17 of the 19 baseline successes (89.5\%). Two previously successful episodes become failures (10.5\% of baseline successes). Among the baseline failure categories, recovery is 9/17 for planning (52.9\%), 12/21 for navigation/base positioning (57.1\%), and 9/43 for manipulation (20.9\%). These are conditional on the baseline episode label, not on entering a particular execution stage. They therefore do not estimate grasp or placement success, nor do they assign recovery to a single tool.

\par\endgroup

\FloatBarrier
\begin{figuretextblock}
\section{Fine-Grained Failure Attribution of the Full System}
\label{app:full_failure_attribution}

Among the 53 failed full-system episodes, 47 are attributed to manipulation, five to navigation and base positioning, and one to task planning. The remaining manipulation failures are mainly incorrect object placement, object drops during transport, and collision-induced scene disruption. This distribution is consistent with the bottleneck shift discussed in the main paper. 

\begin{figure}[H]
    \centering
    \includegraphics[width=0.94\linewidth]{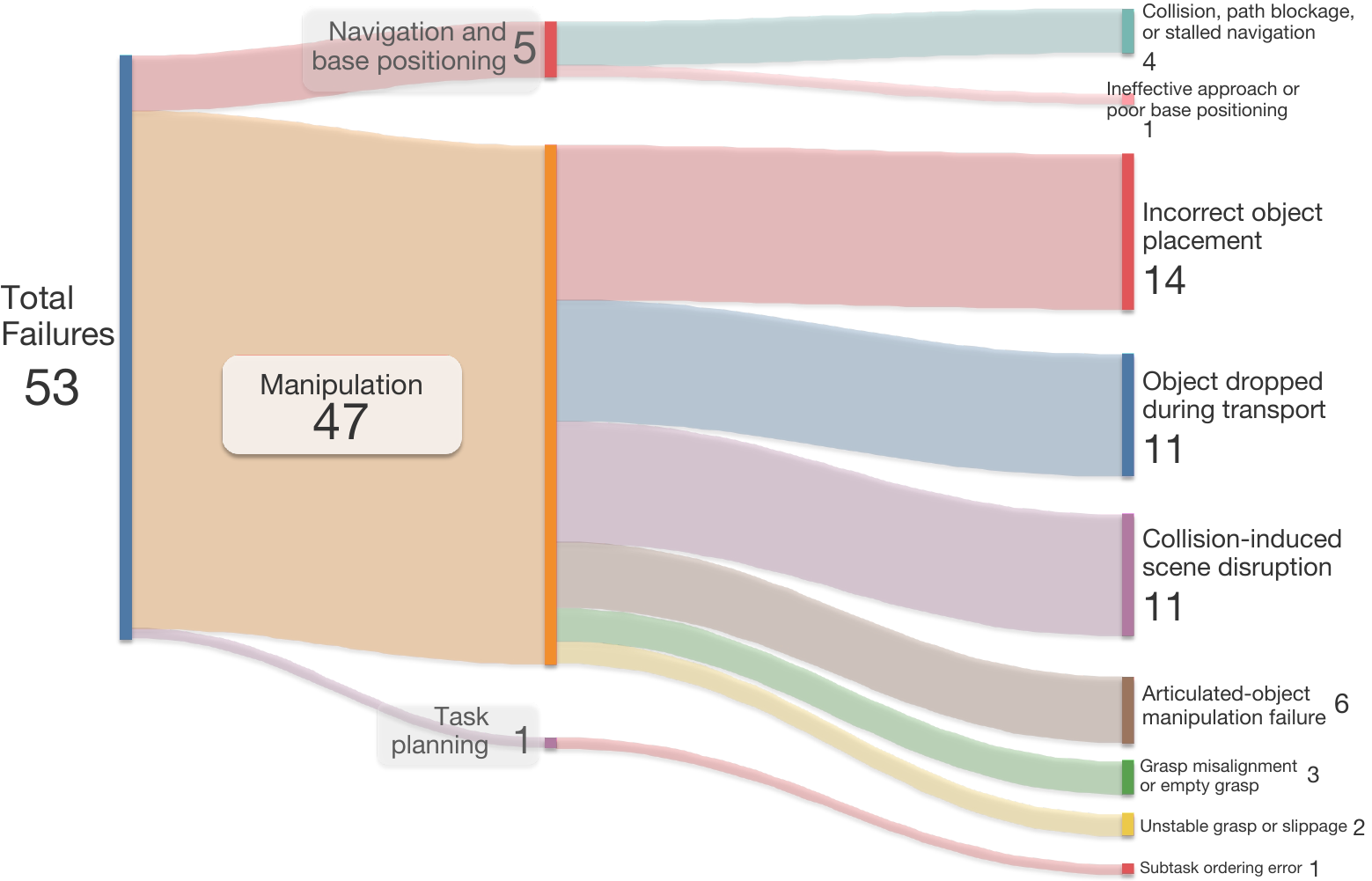}
    
    \caption{
    \textbf{Fine-grained failure attribution for the full system.} Counts are reported over the same 100 matched development episodes.
    }
    \label{fig:full_failure_attribution}

\end{figure}

\end{figuretextblock}

\FloatBarrier
\begingroup
\section{Per-Task and Horizon Results}
\label{app:per_task}

Table~\ref{tab:per_task_combined} combines the task list, official subtask counts, and full-system results on the development episodes. Each task is evaluated with the same five episode seeds used in the main evaluation.

\begin{table}[!htbp]
    \centering
    \caption{\textbf{Per-task results on the 20 mobile manipulation tasks of RoboCasa365.}
    The number of subtasks follows the official RoboCasa365 decomposition. Each task is evaluated with five matched development episode seeds.}
    \label{tab:per_task_combined}
    \small
    \LayoutFit{
    \begin{tabular}{lcc@{\hspace{1.0em}}lcc}
        \toprule
        Task & \# Subtasks & Full & Task & \# Subtasks & Full \\
        \midrule
        DeliverStraw            & 4  & 1/5 & ArrangeBreadBasket       & 5 & 2/5 \\
        GetToastedBread         & 4  & 2/5 & BreadSelection            & 2 & 3/5 \\
        PackIdenticalLunches    & 15 & 1/5 & GarnishPancake            & 4 & 4/5 \\
        ScrubCuttingBoard       & 2  & 4/5 & GatherTableware           & 4 & 2/5 \\
        SearingMeat             & 3  & 2/5 & HeatKebabSandwich         & 6 & 0/5 \\
        SetUpCuttingStation     & 2  & 2/5 & MakeIceLemonade           & 5 & 3/5 \\
        StackBowlsCabinet       & 2  & 5/5 & PortionHotDogs            & 4 & 4/5 \\
        SteamInMicrowave        & 6  & 3/5 & RecycleBottlesByType      & 3 & 4/5 \\
        StirVegetables          & 4  & 1/5 & SeparateFreezerRack       & 7 & 2/5 \\
        StoreLeftoversInBowl    & 5  & 1/5 & WaffleReheat              & 4 & 1/5 \\
        \midrule
        \multicolumn{6}{c}{Overall: 47/100 (47.0\%)} \\
        \bottomrule
    \end{tabular}}
\end{table}
\par\endgroup

We also group tasks by the number of annotated subtasks to provide a descriptive view of performance across task horizons. The full system improves over the baseline in both groups. Observed success is lower for tasks with 5--15 subtasks than for tasks with 2--4 subtasks in both systems.

\begin{table}[H]
    \centering
    \caption{\textbf{Performance across task horizons.}
    Tasks are grouped by the number of annotated subtasks.}
    \label{tab:horizon}
    \small
    \begin{tabular}{lcccc}
        \toprule
        Task horizon & \# Tasks & Baseline & Full system & $\Delta$ (pp) \\
        \midrule
        2--4 subtasks  & 13 & 15/65 (23.1\%) & 35/65 (53.8\%) & +30.7 \\
        5--15 subtasks & 7  & 4/35 (11.4\%)  & 12/35 (34.3\%) & +22.9 \\
        \midrule
        Overall       & 20 & 19/100 (19.0\%) & 47/100 (47.0\%) & +28.0 \\
        \bottomrule
    \end{tabular}
\end{table}

\paragraph{Reference from error compounding.}
The subtask VLA achieves 80\% success on the atomic-task evaluation. As a simple reference for the effect of task length, we treat each annotated subtask as an independent event with success probability $p=0.8$. For a task with $H_i$ subtasks, the reference success rate is $0.8^{H_i}$. Averaging over the 20 tasks gives

\[
\frac{1}{20}\sum_{i=1}^{20}0.8^{H_i}=41.0\%.
\]

The corresponding averages are 49.6\% for the 13 tasks with 2--4 subtasks and 25.0\% for the 7 tasks with 5--15 subtasks. The observed agentic baseline achieves 19.0\%.

\FloatBarrier
\begin{figuretextblock}
\section{Cross-Benchmark Evaluation on BEHAVIOR-1K}
\label{app:behavior1k}

To evaluate the system design beyond RoboCasa365, we test the framework on five BEHAVIOR-1K~\citep{li2023behavior} 2026 Challenge tasks: Pick Up Trash, Make Rose Centerpieces, Composting Waste, Installing Smoke Detectors, and Make Microwave Popcorn.

\begin{table}[H]
    \centering
    \caption{
    Cross-benchmark evaluation on five BEHAVIOR-1K 2026 Challenge tasks. Each method is evaluated on 20 matched episodes per task, using the same scene seeds and initial states across methods. Overall results are aggregated over 100 episodes per method.
    }
    
    \label{tab:behavior1k}
    \small
    \setlength{\tabcolsep}{5pt}
    \LayoutFit{%
    \begin{tabular}{lcccccc}
        \toprule
        Method
        & \shortstack{Pick Up\\Trash}
        & \shortstack{Make Rose\\Centerpieces}
        & \shortstack{Composting\\Waste}
        & \shortstack{Installing Smoke\\Detectors}
        & \shortstack{Make Microwave\\Popcorn}
        & Overall \\
        \midrule

        $\pi_{0.5}$
        & 1 (5\%)
        & 2 (10\%)
        & 1 (5\%)
        & 1 (5\%)
        & 2 (10\%)
        & 7 / 100 (7.0\%) \\

        Harness VLA
        & 2 (10\%)
        & 3 (15\%)
        & 3 (15\%)
        & 2 (10\%)
        & 4 (20\%)
        & 14 / 100 (14.0\%) \\

        Ours
        & \textbf{7 (35\%)}
        & \textbf{9 (45\%)}
        & \textbf{7 (35\%)}
        & \textbf{6 (30\%)}
        & \textbf{8 (40\%)}
        & \textbf{37 / 100 (37.0\%)} \\

        \bottomrule
    \end{tabular}%
    }
\end{table}
\end{figuretextblock}

For this evaluation, we train the low-level policies using only the training data provided by the BEHAVIOR-1K challenge. Following the RoboCasa365 setup, our subtask VLA is trained from trajectories with the provided subtask annotations: each trajectory is segmented according to its subtask boundaries and conditioned on the corresponding subtask instruction. In contrast, the $\pi_{0.5}$ baseline is trained on the corresponding complete trajectories without subtask-level supervision. For the agentic methods, Kimi K3 is used as the high-level planner. Each method is evaluated for 20 rollouts per task, resulting in 100 rollouts in total. Corresponding rollouts across methods use the same scene seed and identical initial simulator and object states.

As shown in Table~\ref{tab:behavior1k}, the full system consistently improves over Harness VLA across all five tasks, increasing the overall success rate from 14.0\% to 37.0\%, while the end-to-end model $\pi_{0.5}$ achieves 7.0\%. This experiment evaluates transfer of the agentic system design, with the low-level policies retrained on BEHAVIOR-1K training data. 


\FloatBarrier
\section{Real-Robot Evaluation}
\label{app:real_robot}

We further evaluate the framework on Astribot-S1, a wheeled mobile dual-arm robot equipped with parallel grippers. We consider two real-robot tasks, \textit{Storage Box Transport} and \textit{Table Service}.

\paragraph{Tasks.}
For \textit{Storage Box Transport}, the robot starts away from a cabinet and must move a storage box from the cabinet to a table. The high-level task instruction is:

\begin{quote}
\emph{Move the storage box from the cabinet to the table.}
\end{quote}

The task contains three main subtasks:
\begin{enumerate}
    \item Go to the cabinet and pick up the storage box.
    \item Move to the table with the storage box.
    \item Place the storage box on the table.
\end{enumerate}

For \textit{Table Service}, the robot serves a prepared meal from a serving cart to a dining table. The high-level task instruction is:

\begin{quote}
\emph{Serve the meal at the dining table by setting out the prepared dish, adding the topping, and placing the wine glass beside it.}
\end{quote}

The task contains four main subtasks:
\begin{enumerate}
    \item Move the prepared dish from the serving cart to the dining table.
    \item Add the topping from the bowl to the prepared dish.
    \item Place the bowl on the serving cart.
    \item Pick up the wine glass and place it beside the dish.
\end{enumerate}

Storage Box Transport requires navigation before pickup, object transport during navigation, and placement at a second workspace. Table Service contains multiple manipulation stages.

\paragraph{VLA training and agent setup.}
For both tasks, we initialize the VLA from the released $\pi_{0.5}$~\citep{black2025pi05} checkpoint and fine-tune it on real-robot demonstrations. We collect 500 demonstrations for Storage Box Transport and 300 demonstrations for Table Service.

The end-to-end VLA and the agentic methods use different forms of supervision from the same real-robot demonstrations. For the end-to-end VLA baseline, $\pi_{0.5}$ is fine-tuned directly on complete trajectories conditioned on the high-level task instruction. For the agentic methods, we first use a VLM (Gemini 3.5 Flash~\citep{google2026gemini35flash}) to convert the demonstrations into subtask-level training data. Each segment is paired with a subtask instruction, and $\pi_{0.5}$ is then fine-tuned on these subtask-level examples. The resulting policy is used as the subtask VLA for both the agentic baseline and the full system. Kimi K3 serves as the high-level planner for both agentic methods, following the simulation experiments. The planner selects a subtask instruction for the subtask VLA based on the current observation and execution history. The agentic baseline uses the planner together with the subtask VLA. The full system further uses demonstration-based planning guidance, adaptive visual memory, base adjustment, and end-effector adjustment.

\Needspace*{16\baselineskip}
\paragraph{Evaluation.}
We compare three settings: the end-to-end $\pi_{0.5}$ VLA, the agentic baseline, and the full system. Each method is evaluated over 20 real-robot trials. 

\begin{table}[H]
    \centering
    \caption{
        \textbf{Real-robot evaluation.} Each method is evaluated for 10 trials on each task.
    }
    \label{tab:real_robot_results}
    \small
    \setlength{\tabcolsep}{5pt}
    \begin{tabular}{lccc}
        \toprule
        Method
        & Storage Box Transport
        & Table Service
        & Overall \\
        \midrule
        $\pi_{0.5}$ (end-to-end VLA)
        & 2/10
        & 2/10
        & 4/20 \\

        Agentic baseline
        & 2/10
        & 3/10
        & 5/20 \\

        Full system
        & \textbf{6/10}
        & \textbf{7/10}
        & \textbf{13/20} \\
        \bottomrule
    \end{tabular}
\end{table}

Across the two tasks, the full system succeeds in 13/20 trials, compared with 5/20 for the agentic baseline. The improvement is observed on both tasks rather than being concentrated in a single task. These results follow the same overall trend as the simulation evaluation, where the added interventions improve long-horizon task completion around the same agentic structure.

\begin{figuretextblock}
\paragraph{Qualitative example.}
Figure~\ref{fig:real_robot_rollout} shows a representative successful rollout of the Table Service task. The robot first transfers the prepared dish from the serving cart to the dining table, adds the topping, and then places the wine glass beside the dish.

\begin{figure}[H]
    \centering
    \includegraphics[width=0.92\linewidth]{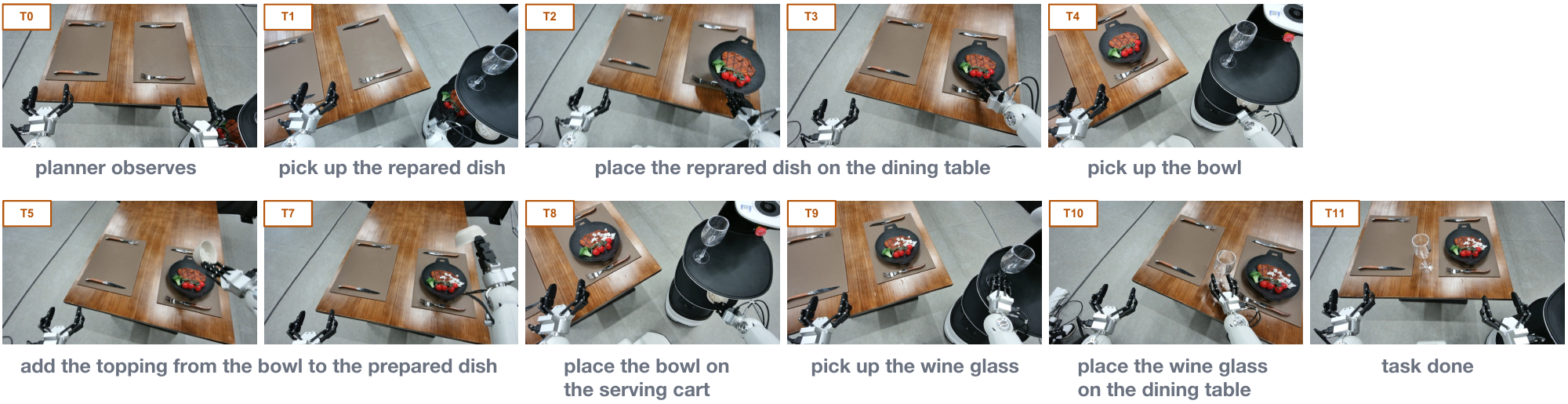}
    \caption{
        \textbf{Real-robot execution of the Table Service task.}
        A representative successful rollout of the full system on Astribot-S1. The robot moves the prepared dish from the serving cart to the dining table, adds the topping, and places the wine glass beside the dish.
    }
    \label{fig:real_robot_rollout}
\end{figure}
\end{figuretextblock}

\FloatBarrier
\section{Leave-One-Out and Design Ablations}
\label{app:loo}

The progressive integration in Figure \ref{fig:progressive} follows the order motivated by the baseline failure analysis, so the incremental gain of a component can depend on which components are already present. We therefore perform a leave-one-out ablation from the full system. Each configuration removes one intervention group while keeping all other components unchanged and uses the same 100 matched development episodes.

\begin{table}[H]
    \centering
    \caption{\textbf{Leave-one-out ablation of intervention groups on the development episodes.}}
    \label{tab:loo_dev}
    \small
    \begin{tabular}{lcc}
        \toprule
        Configuration & Task Success Rate(\%) & $\Delta$ from Full (pp) \\
        \midrule
        Full system                         & \textbf{47.0} & -- \\
        $-$ Skill library                   & 39.0 & -8.0 \\
        $-$ Adaptive visual memory          & 41.0 & -6.0 \\
        $-$ Base adjustment                 & 35.0 & \textbf{-12.0} \\
        $-$ End-effector adjustment         & 40.0 & -7.0 \\
        \bottomrule
    \end{tabular}
\end{table}

Removing base adjustment produces the largest observed drop, from 47.0\% to 35.0\%, consistent with its large gain in the progressive experiment. Removing the other intervention groups also reduces the success rate. Paired confidence intervals and McNemar tests for these comparisons, together with the held-out leave-one-out results, are reported in Appendix~\ref{app:stats}.

\begin{figuretextblock}
\subsection{Fine-Grained Ablations}
\label{app:i_4}

The fine-grained ablations in Table~2 also use matched episode seeds and simulator initial states.
Table~\ref{tab:paired_fine} reports the paired comparisons.

\begin{table}[H]
    \centering
    \caption{\textbf{Paired analysis of the fine-grained ablations in Table~2.}
    All comparisons use the same 100 matched development episodes.}
    \label{tab:paired_fine}
    \small
    \LayoutFit{
    \begin{tabular}{lcccccc}
        \toprule
        Comparison & Success (\%) & $\Delta$ (pp) & $n_{01}$ & $n_{10}$ & 95\% CI (pp) & McNemar $p$ \\
        \midrule
        \multicolumn{7}{l}{\textit{Skill library}} \\
        HIW-500 $\rightarrow$ in-domain video
            & 40 $\rightarrow$ 47 & +7 & 12 & 5 & [-1, 15] & 0.143 \\
        Text-only $\rightarrow$ full skill card
            & 43 $\rightarrow$ 47 & +4 & 7 & 3 & [-2, 10] & 0.344 \\
        \midrule
        \multicolumn{7}{l}{\textit{Adaptive visual memory}} \\
        Uniform-8 $\rightarrow$ Adaptive-8
            & 37 $\rightarrow$ 42 & +5 & 8 & 3 & [-1, 11] & 0.227 \\
        Uniform-16 $\rightarrow$ Adaptive-16
            & 40 $\rightarrow$ 47 & +7 & 10 & 3 & [0, 14] & 0.092 \\
        Uniform-32 $\rightarrow$ Adaptive-32
            & 35 $\rightarrow$ 45 & +10 & 13 & 3 & [2, 18] & 0.021 \\
        Adaptive-8 $\rightarrow$ Adaptive-16
            & 42 $\rightarrow$ 47 & +5 & 9 & 4 & [-2, 12] & 0.267 \\
        Adaptive-32 $\rightarrow$ Adaptive-16
            & 45 $\rightarrow$ 47 & +2 & 6 & 4 & [-4, 8] & 0.754 \\
        \midrule
        \multicolumn{7}{l}{\textit{Base adjustment}} \\
        w/o \texttt{check\_stance} $\rightarrow$ Full
            & 36 $\rightarrow$ 47 & +11 & 14 & 3 & [3, 19] & 0.013 \\
        w/o \texttt{approach\_xy} $\rightarrow$ Full
            & 42 $\rightarrow$ 47 & +5 & 10 & 5 & [-2, 12] & 0.302 \\
        w/o \texttt{goto\_stance} $\rightarrow$ Full
            & 39 $\rightarrow$ 47 & +8 & 12 & 4 & [0, 16] & 0.077 \\
        \midrule
        \multicolumn{7}{l}{\textit{End-effector adjustment}} \\
        Planner target $\rightarrow$ geometric target
            & 43 $\rightarrow$ 47 & +4 & 9 & 5 & [-3, 11] & 0.424 \\
        \bottomrule
    \end{tabular}}
\end{table}
\end{figuretextblock}

The paired results are consistent with the descriptive ablations in the main paper. Adaptive memory has higher observed success than the corresponding uniform-history setting at all three frame budgets, although the uncertainty remains wide for several comparisons. Within base adjustment, removing the accessibility check produces the clearest drop. The difference between planner-specified and geometrically estimated end-effector targets is smaller and its confidence interval includes zero.

\begingroup
\subsection{Separating Memory Access and Frame Selection}
\label{sec:memory_factor}

The visual-memory ablation in Table~\ref{tab:agentic_ablation} changes two factors at once: when memory is accessed and which history frames are returned. We separate these factors using the same 16-frame budget. Memory is either provided at every planner turn or accessed only when requested by the planner. Within each setting, frames are either uniformly sampled from the episode history or ranked by their relevance to the retrieval query. All four configurations use the same 100 matched development episodes.

For planner-triggered uniform memory, the planner decides when memory is needed, but the returned frames are uniformly sampled from the available episode history. For every-turn relevance-based memory, retrieval is performed at every planner turn, using a query generated from the current task state and execution history to rank past observations. All other memory settings, including the candidate history, frame budget, and returned trajectory summary, are kept unchanged.

\begin{table}[H]
    \centering
    \caption{
        \textbf{Separating memory access and frame selection.} All configurations use a 16-frame memory budget and the same 100 matched development episodes.
    }
    \label{tab:memory_factorized}
    \small
    \setlength{\tabcolsep}{5pt}
    \begin{tabular}{lcc}
        \toprule
        Memory Access
        & Uniform
        & Relevance-based \\
        \midrule
        Every turn
        & 40.0
        & 45.0 \\
        Planner-triggered
        & 44.0
        & \textbf{47.0} \\
        \bottomrule
    \end{tabular}
\end{table}

Relevance-based frame selection improves the observed success rate under both access settings, while planner-triggered access also improves over providing memory at every turn. Combining planner-triggered access with relevance-based retrieval gives the highest observed success rate. These results suggest that both when memory is accessed and which observations are retrieved contribute to the adaptive-memory design.

\par\endgroup

\FloatBarrier
\pagebreak[4]
\begingroup
\setlength{\intextsep}{3pt plus 1pt minus 1pt}
\noindent\begin{minipage}[t][\dimexpr\textheight-7pt\relax][s]{\linewidth}
\section{Qualitative Execution Examples}
\label{app:qualitative}

We provide four representative rollouts to illustrate how the high-level planner composes the
fixed subtask VLA with the system-level interventions during long-horizon execution. The examples
cover standard multi-stage execution, coordination between navigation and manipulation,
accessibility-aware base adjustment, and local end-effector correction after a failed placement.

\begin{figure}[H]
    \centering
    \includegraphics[width=0.755\linewidth]{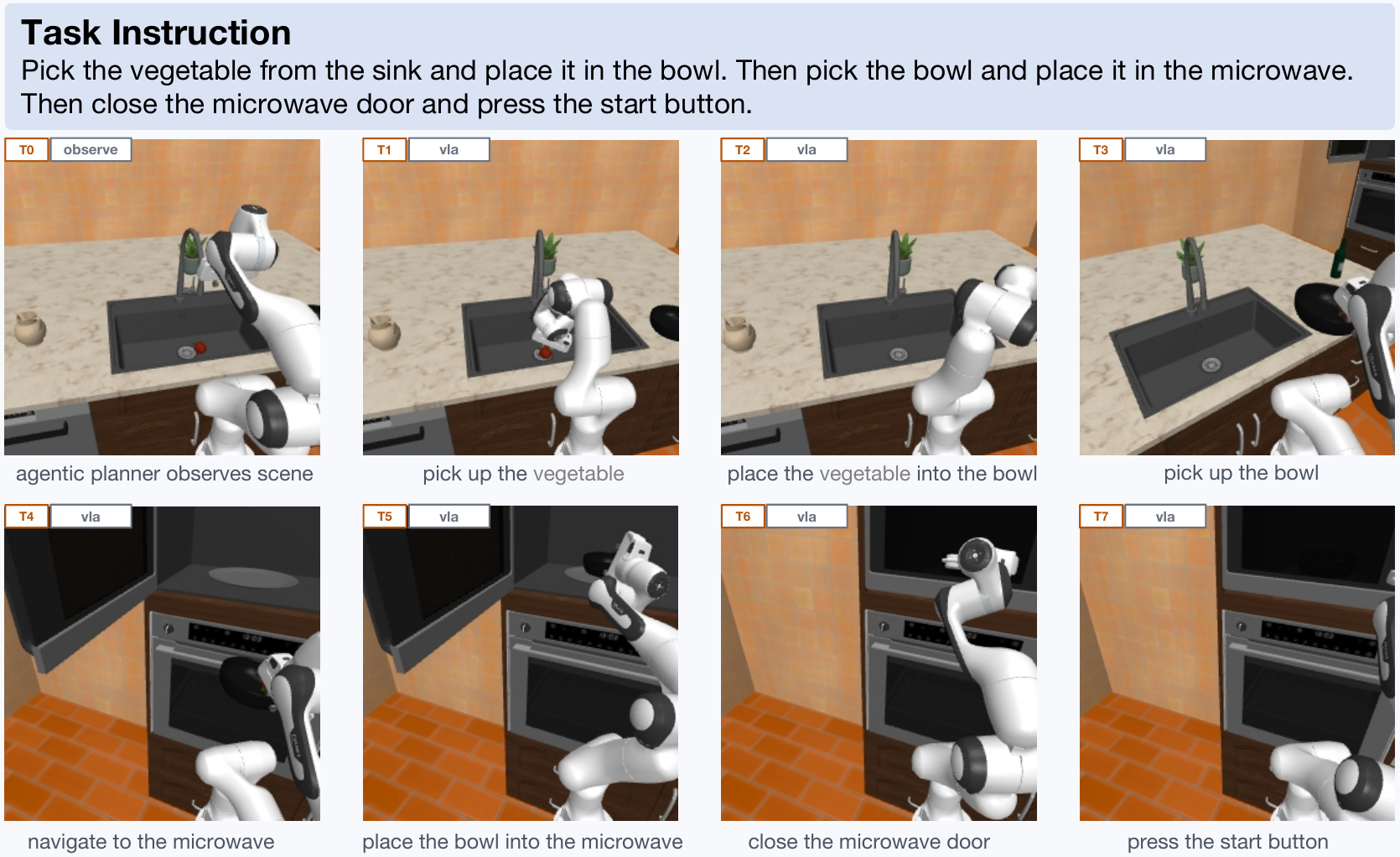}
    
    \caption{
    \textbf{Multi-stage closed-loop execution.}
    The high-level planner decomposes the task into a sequence of subtask-VLA calls, including
    object pickup and placement, navigation to the microwave, interaction with the microwave door,
    and pressing the start button.
    }
    \label{fig:1}

\end{figure}

\vfill
\begin{figure}[H]
    \centering
    \includegraphics[width=0.755\linewidth]{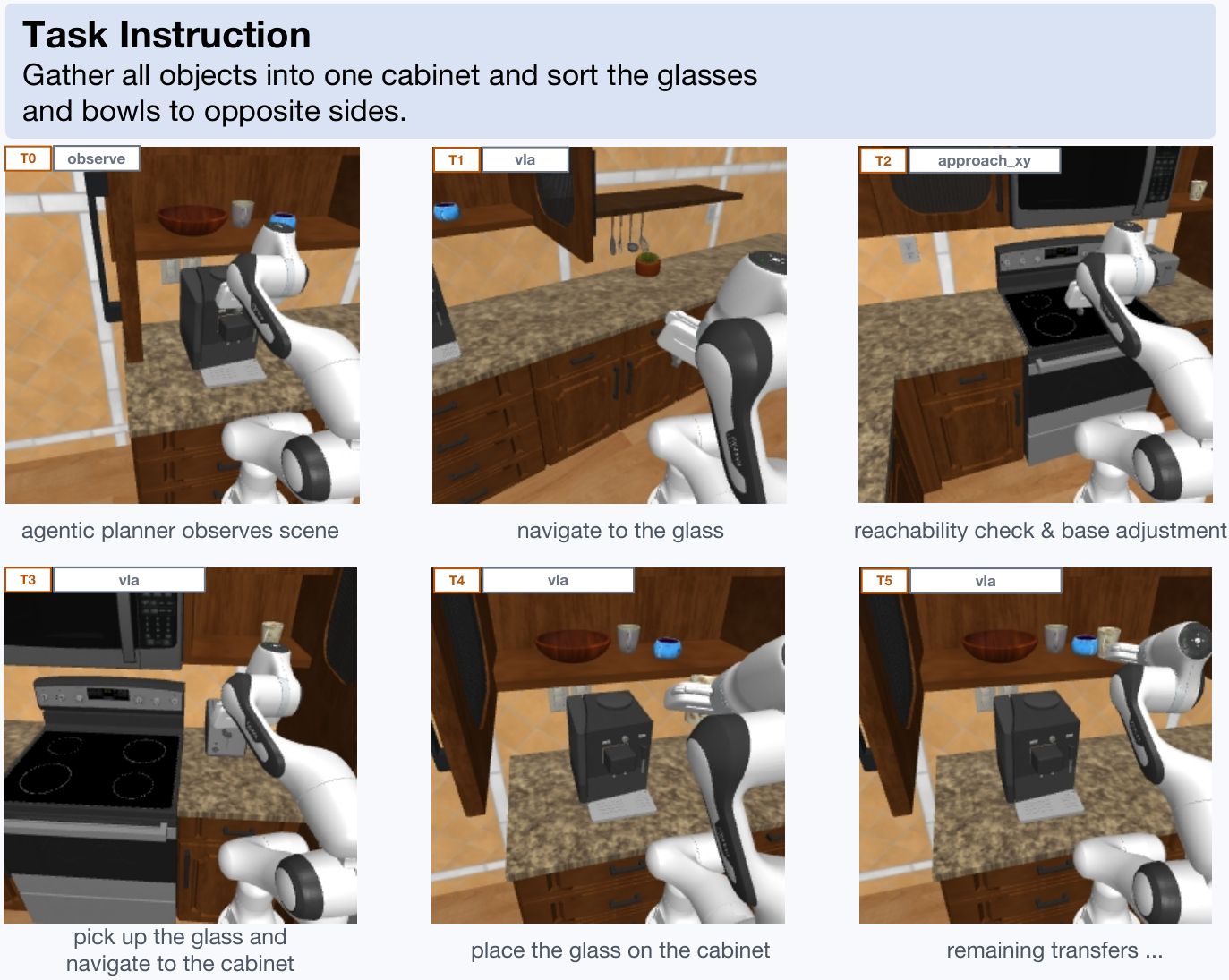}
    
    \caption{
    \textbf{Accessibility-aware base adjustment before manipulation.}
    After navigating near the target, the planner checks whether the current base pose is suitable
    for manipulation and invokes \texttt{approach\_xy} for a local base adjustment before continuing
    with the subtask VLA.
    }
    \label{fig:2}

\end{figure}

\end{minipage}\par
\newpage
\noindent\begin{minipage}[t][\dimexpr\textheight-7pt\relax][s]{\linewidth}
\begin{figure}[H]
    \centering
    \includegraphics[width=0.92\linewidth]{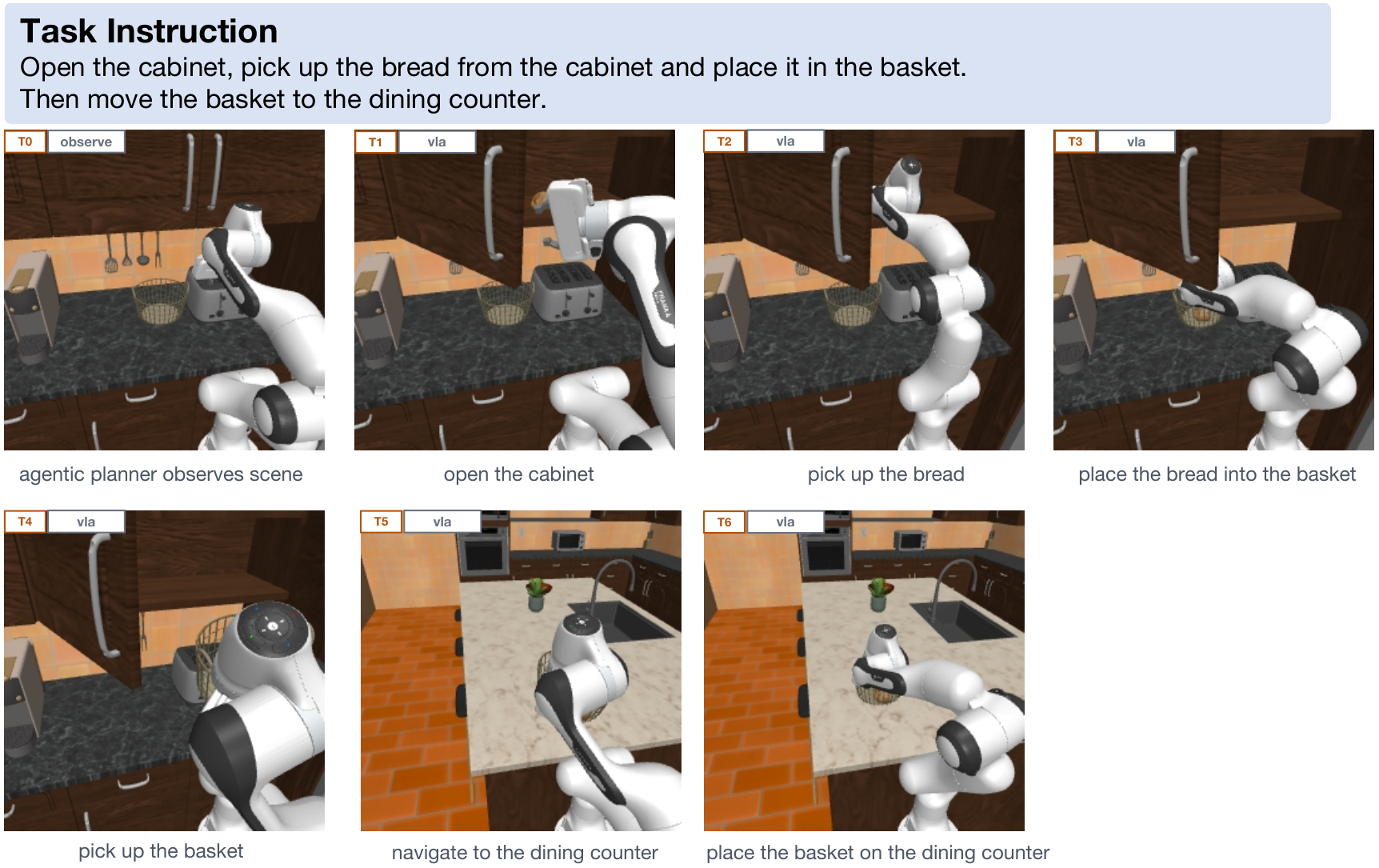}
    
    \caption{
    \textbf{Coordinated navigation and manipulation in a long-horizon task.}
    The planner uses the same subtask VLA to open the cabinet, pick and place the bread, transport
    the basket, navigate to the dining counter, and complete the final placement.
    }
    \label{fig:3}

\end{figure}

\vfill
\begin{figure}[H]
    \centering
    \includegraphics[width=0.92\linewidth]{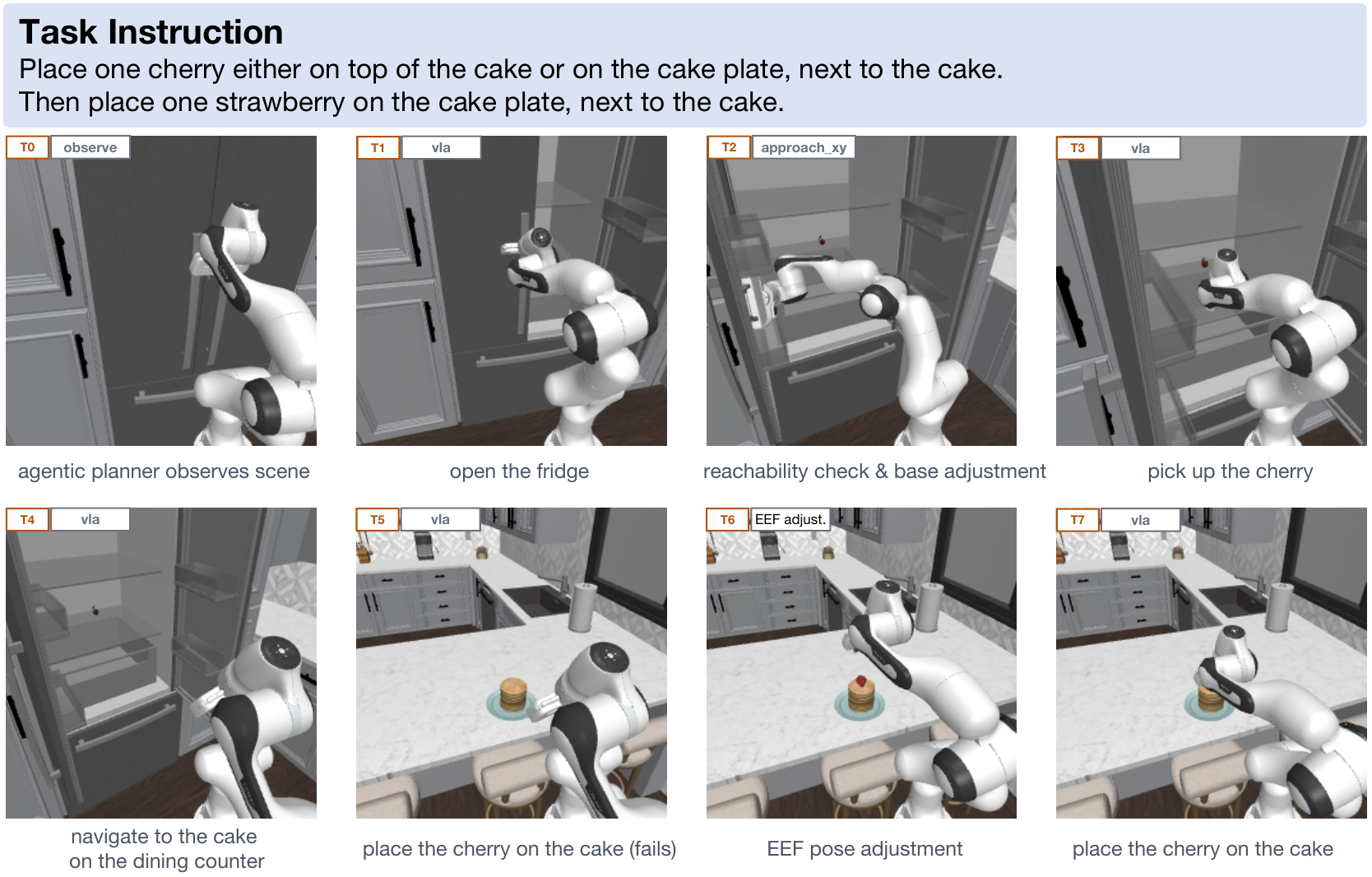}
    
    \caption{
    \textbf{End-effector adjustment after a failed placement attempt.}
    After the initial placement fails, the planner invokes a local end-effector adjustment and
    retries the placement with the same subtask VLA.
    }
    \label{fig:4}

\end{figure}

\end{minipage}\par

\endgroup

\FloatBarrier
\endgroup

\end{document}